\documentclass[runningheads]{llncs}
\usepackage{graphicx}
\usepackage{comment}
\usepackage{amsmath,amssymb} 
\usepackage{color}
\usepackage{epsfig}
\usepackage{blindtext, rotating}
\usepackage{array}
\usepackage{booktabs}
\usepackage{multirow}
\usepackage{hyperref}
\usepackage{bbm}
\usepackage{flushend}
\usepackage{animate}
\usepackage[nocompress]{cite}
\usepackage{pifont}
\usepackage[utf8]{inputenc}
\usepackage[T1]{fontenc}
\newcommand{\cmark}{\text{\ding{51}}}
\newcommand{\xmark}{\text{\ding{55}}}

\begin{document}
\pagestyle{headings}
\mainmatter
\def\ECCVSubNumber{1586}  

\title{Generating Videos of Zero-Shot Compositions of Actions and Objects} 

\titlerunning{Generating Videos of Zero-Shot Compositions of Actions and Objects}
%
\author{Megha Nawhal\inst{1,2}
\and
Mengyao Zhai\inst{2}
\and
Andreas Lehrmann\inst{1} 
\and
Leonid Sigal\inst{1,3,4,5}
\and
Greg Mori\inst{1,2}
}
\authorrunning{M. Nawhal et al.}

\institute{Borealis AI, Vancouver, Canada
\and
Simon Fraser University,
Burnaby, Canada
\and
University of British Columbia,
Vancouver, Canada
\and
Vector Institute for AI \quad $^{5}$ CIFAR AI Chair
}
\maketitle

\begin{abstract}
Human activity videos involve rich, varied interactions between people and objects. In this paper we develop methods for generating such videos -- making progress toward addressing the important, open problem of video generation in complex scenes.  In particular, we introduce the task of generating human-object interaction videos in a zero-shot compositional setting, \textit{i.e.},  generating videos for action-object compositions that are unseen during training, having seen the target action and target object separately. This setting is particularly important for generalization in human activity video generation, obviating the need to observe every possible action-object combination in training and thus avoiding the combinatorial explosion involved in modeling complex scenes. To generate human-object interaction videos, we propose a novel adversarial framework HOI-GAN which includes multiple discriminators focusing on different aspects of a video. To demonstrate the effectiveness of our proposed framework, we perform extensive quantitative and qualitative evaluation on two challenging datasets: EPIC-Kitchens and 20BN-Something-Something v2.
\end{abstract}

\section{Introduction}

Visual imagination and prediction are fundamental components of human intelligence. Arguably, the ability to create realistic renderings from symbolic representations are considered prerequisite for broad visual understanding.
%
Computer vision has seen rapid advances in the field of image generation over the past few years.  Existing models are capable of generating impressive results in this static scenario, ranging from hand-written digits~\cite{goodfellow2014generative,denton2015deep,arjovsky2017wasserstein} to realistic scenes~\cite{van2016conditional,zhang2017stackgan,karras2017progressive,brock2018large,isola2017image}.  Progress on \emph{video generation} \cite{vondrick2016generating,saito2017temporal,tulyakov2018mocogan,he2018probabilistic,wang2018vid2vid,bansal2018recycle,wang2019fewshotvid2vid}, on the other hand, has been relatively moderate and remains an open and challenging problem. While most approaches focus on the expressivity and controllability of the underlying generative models, their ability to generalize to unseen scene compositions has not received as much attention. However, such generalizability is an important cornerstone of robust visual imagination as it demonstrates the capacity to reason over elements of a scene. 

\begin{figure}[t!]
\centering
\includegraphics[width=0.5\linewidth]{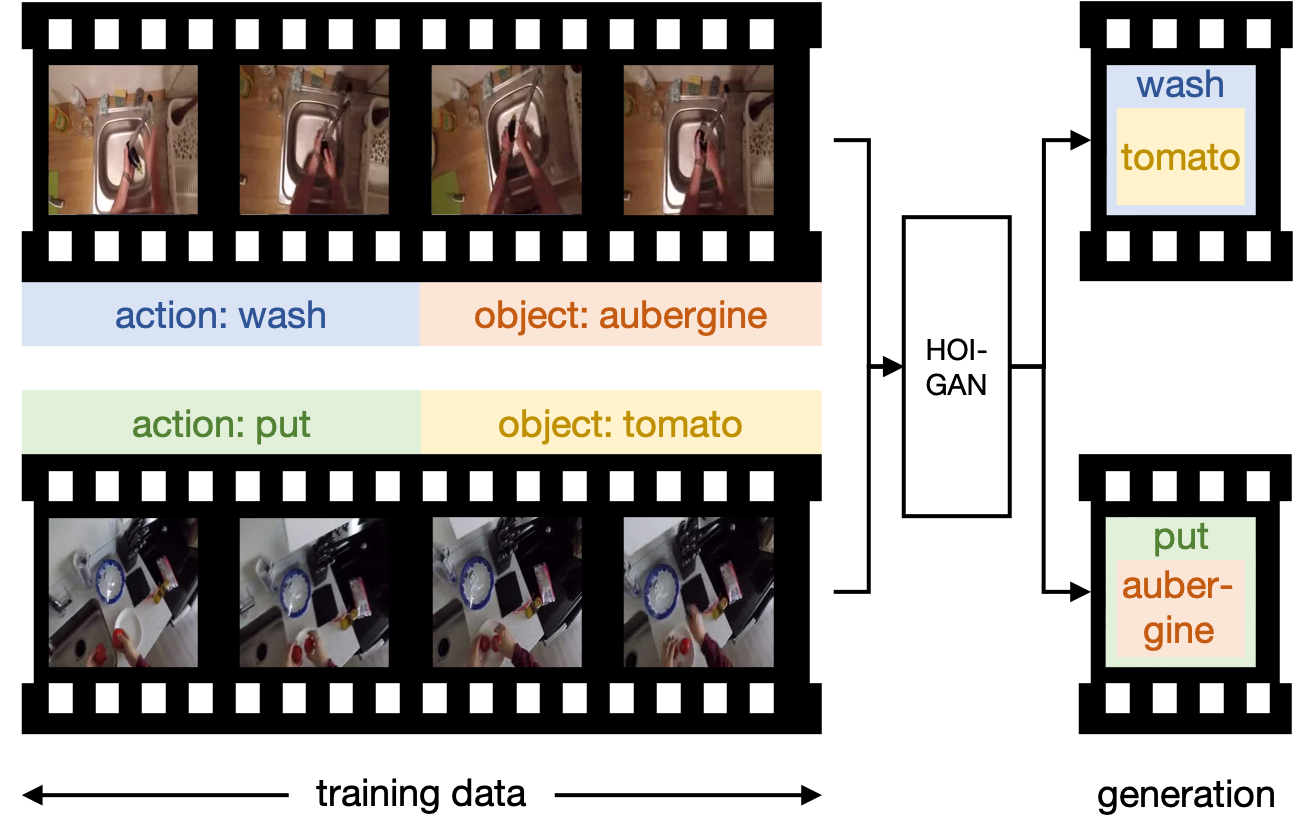}
\caption{{\textbf{Generation of Zero-Shot Human-Object Interactions.}} Given training examples \emph{``wash aubergine"} and \emph{``put tomato"}, an intelligent agent should be able to 
imagine action sequences for unseen action-object compositions, \textit{i.e.}, \emph{``wash tomato"} and \emph{``put aubergine".}}
\label{fig:teaser}
\end{figure}

We posit that the domain of human activities constitutes a rich realistic testbed for video generation models. Human activities involve people interacting with objects in complex ways, presenting numerous challenges for generation -- the need to (1) render a variety of objects; (2) model the temporal evolution of the effect of actions on objects; (3) understand spatial relations and interactions; and (4) overcome the paucity of data for the complete set of action-object pairings. 
The last, in particular, is a critical challenge that also serves as an opportunity for designing and evaluating generative models that can generalize to myriad, possibly unseen, action-object compositions. 
For example, consider
Figure~\ref{fig:teaser}. The activity sequences for \emph{``wash aubergine"} (action $a_1$: ``wash"; object $o_1$: ``aubergine") and \emph{``put tomato"}(action $a_2$: ``put"; object $o_2$: ``tomato") are observed in the training data. A robust visual imagination would then allow an agent to imagine videos for \emph{``wash tomato"} $(a_1,o_2)$ and \emph{``put aubergine"} $(a_2,o_1)$. 

We propose a novel framework for generating human-object interaction (HOI) videos for unseen action-object compositions. We refer to this task as \emph{zero-shot HOI video generation}. To the best of our knowledge, our work is the first to propose and address this problem. In doing so, we push the envelope on conditional (or controllable) video generation and focus squarely on the model's ability to generalize to unseen action-object compositions. This zero-shot compositional setting verifies that the model is capable of semantic disentanglement of the action and objects in a given context and recreating them separately in other contexts.   

The desiderata for performing zero-shot HOI video generation include: (1) mapping the content in the video to the right semantic category, (2) ensuring spatial and temporal consistency across the frames of a video, and (3) producing interactions with the right object in the presence of multiple objects. Based on these observations, we introduce a novel multi-adversarial learning scheme involving multiple discriminators, each focusing on different aspects of an HOI video. Our framework \textit{HOI-GAN} generates a fixed length video clip given an action, an object, and a target scene serving as the context. 

Concretely, the conditional inputs to our framework are semantic labels of action and object, and a single start frame with a mask providing the background and location for the object. Then, the model has to create the object, reason over the action, and enact the action on the object (leading to object translation and/or transformation) over the background, thus generating the whole interaction video. During training of the generator, our framework utilizes four discriminators -- three pixel-centric discriminators, namely, {\em frame} discriminator, {\em gradient} discriminator, {\em  video} discriminator; and one object-centric {\em relational} discriminator. The three pixel-centric discriminators ensure spatial and temporal consistency across the frames. The novel relational discriminator leverages spatio-temporal scene graphs to reason over the object layouts in videos ensuring the right interactions among objects. Through experiments, we show that our HOI-GAN framework is able to disentangle objects and actions and learns to generate videos with unseen 
compositions.

In summary, our contributions are as follows:
\begin{itemize}
    \item We introduce the task of zero-shot HOI video generation. Specifically, given a training set of videos depicting certain action-object compositions, we propose to generate unseen compositions having seen the target action and target object individually, \textit{i.e.}, the target action was paired with a different object and the target object was involved in a different action. 
    \item We propose a novel adversarial learning scheme and introduce our HOI-GAN framework to generate HOI videos in a zero-shot compositional setting.
    \item We demonstrate the effectiveness of HOI-GAN through empirical evaluation on two challenging HOI video datasets: \emph{20BN-something-something v2}\cite{goyal2017something} and \emph{EPIC-Kitchens} \cite{Damen2018EPICKITCHENS}. We perform both quantitative and qualitative evaluation of the proposed approach and compare with state-of-the-art approaches.
\end{itemize}
Overall, our work facilitates research in the direction of enhancing generalizability of generative models for complex videos.

\section{Related Work}
Our paper builds on prior work in: (1) modeling of human-object interactions and (2) GAN-based video generation. In addition, we also discuss literature relevant to HOI video generation in a zero-shot compositional setting. 

\noindent
\textbf{Modeling Human-Object Interactions.}  
Earlier research attempts to study human-object interactions (HOIs) aimed at studying object affordances \cite{grabner2011makes,kjellstrom2011visual} and semantic-driven understanding of object functionalities \cite{stark1991achieving,gupta2007objects}. Recent work on modeling HOIs in images range from studying semantics and spatial features of interactions between humans and objects \cite{delaitre2012scene,zellers2018neural,gkioxari2018detecting} to action information\cite{fouhey2014people,desai2012detecting,yao2010modeling}. Furthermore, there have been attempts to create large scale image and video datasets to study HOI \cite{krishna2017visual,chao2015hico,chao2018learning,goyal2017something}. To model dynamics in HOIs, recent works have proposed methods that jointly model actions and objects in videos \cite{kato2018compositional,sigurdsson2017actions,kalogeiton2017joint}. 
Inspired by these approaches, we model HOI videos as  compositions of actions and objects.

\noindent
\textbf{GAN-based Image \& Video Generation.}  
Generative Adversarial Network (GAN)~\cite{goodfellow2014generative} and its variants~\cite{denton2015deep,arjovsky2017wasserstein,zhao2017energy} have shown tremendous progress in high quality image generation. 
Built over these techniques, conditional image generation using various forms of inputs to the generator such as textual information \cite{reed2016generative,zhang2017stackgan,xu2018attngan}, category labels \cite{odena2017conditional,miyato2018cgans}, and images \cite{kim2017learning,isola2017image,zhu2017unpaired,liu2017unsupervised} have been widely studied. This class of GANs allows the generator network to learn a mapping between conditioning variables and the real data distribution, thereby allowing control over the generation process. 
Extending these efforts to conditional video generation is not straightforward as generating a video involves modeling of both spatial and temporal variations. Vondrick et al. ~\cite{vondrick2016generating} proposed the Video GAN (VGAN) framework to generate videos using a two-stream generator network that decouples foreground and background of a scene. Temporal GAN (TGAN)~\cite{saito2017temporal} employs a separate generator for each frame in a video and an additional generator to model temporal variations across these frames. 
MoCoGAN~\cite{tulyakov2018mocogan} disentangles the latent space representations of motion and content in a video to perform controllable video generation using seen compositions of motion and content as conditional inputs. In our paper, we evaluate the extent to which these video generation methods generalize when provided with unseen scene compositions as conditioning variables. Furthermore, promising success has been achieved by recent video-to-video translation methods~\cite{wang2018vid2vid,wang2019fewshotvid2vid,bansal2018recycle} wherein video generation is conditioned on a corresponding semantic video. In contrast, our task does not require semantic videos as conditional input.

\noindent
\textbf{Video Prediction.} Video prediction approaches predict future frames of a video given one or a few observed frames using RNNs \cite{srivastava2015unsupervised}, variational auto-encoders ~\cite{walker2016uncertain,walker2017pose}, adversarial training~\cite{mathieu2016deep,liang2017dual}, or auto-regressive methods \cite{kalchbrenner2017video}. 
While video prediction is typically posed as an image-conditioned (past frame) image generation (future frame) problem, it is substantially different from video generation where the goal is to generate a video clip given a stochastic latent space. 

\noindent
\textbf{Video Inpainting.} Video inpainting/completion refers to the problem of correctly filling up the missing pixels given a video with arbitrary spatio-temporal pixels missing \cite{newson2014video,shen2006video,granados2012background,ebdelli2015video,niklaus2017video}. In our setting, however, the model only receives a single static image as input and not a video. Our model is required to go beyond merely filling in pixel values and has to produce an output video with the right visual content depicting the prescribed action upon a synthesized object. In doing so, the background may, and in certain cases should, evolve as well. 


\noindent
\textbf{Zero-Shot Learning.} Zero-shot learning (ZSL) aims to solve the problem of recognizing classes whose instances are not seen during training. In ZSL, external information of a certain form is required to share information between classes to transfer knowledge from seen to unseen classes. A variety of techniques have been used for ZSL ranging from usage of attribute-based information~\cite{lampert2009learning,farhadi2009describing}, word embeddings~\cite{xian2018feature} to WordNet hierarchy~\cite{akata2015evaluation} and text-based descriptions~\cite{guadarrama2013youtube2text,elhoseiny2013write,zhu2018generative,lei2015predicting}. \cite{xian2017zero} provides a thorough overview of zero-shot learning techniques. Similar to these works, we leverage word embeddings to reason over the unseen compositions of actions and objects in the context of video generation. 

\noindent
\textbf{Learning Visual Relationships.} Visual relationships in the form of scene graphs, \textit{i.e.}, directed graphs representing relationships (edges) between the objects (nodes) have been used for image caption evaluation~\cite{anderson2016spice}, image retrieval \cite{johnson2015image} and predicting scene compositions for images~\cite{xu2017scene,lu2016visual,newell2017pixels}. 
Furthermore, in a generative setting, \cite{johnson2018image} aims to synthesize an image from a given scene graph and evaluate the generalizability of an adversarial network to create images with unseen relationships between objects. Similarly, we leverage spatio-temporal scene graphs to learn relevant relations among the objects and focus on the generalizability of video generation models to unseen compositions of actions and objects. However, our task of zero-shot HOI video generation is more difficult as it requires learning to map the inputs to spatio-temporal variations in a video.

\noindent
\textbf{Learning Disentangled Representations for Videos.} Various methods have been proposed to learn disentangled representations in videos \cite{tulyakov2018mocogan,hsieh2018learning,denton2017unsupervised}, such as, learning representations by decoupling the content and pose~\cite{denton2017unsupervised}, or separating motion from content using image differences~\cite{villegas2017decomposing}. Similarly, our model implicitly learns to disentangle the action and object information of an HOI video.

\section{HOI-GAN}
\begin{figure}[t]
\centering
         \includegraphics[width=0.47\textwidth]{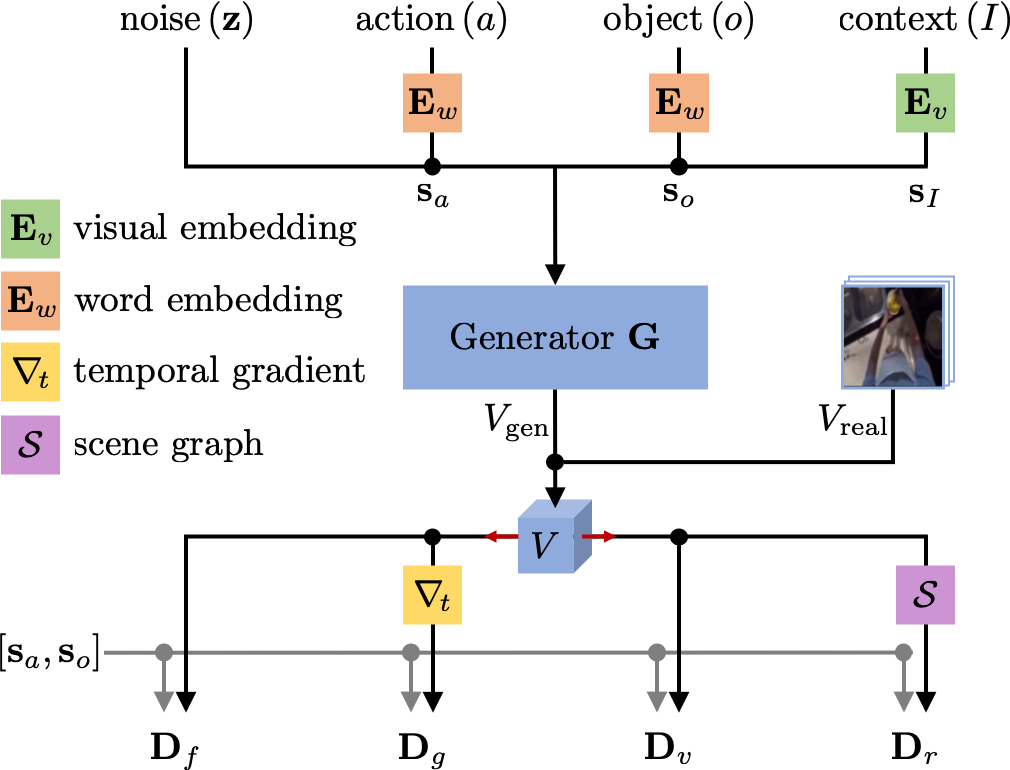}
    \caption{{\bf Architecture Overview.} The generator network $\mathbf{G}$ is trained using 4 discriminators simultaneously: a frame discriminator $\mathbf{D}_f$, a gradient discriminator $\mathbf{D}_g$, a video discriminator $\mathbf{D}_v$, and a relational discriminator $\mathbf{D}_{r}$. Given the word embeddings of an action $\mathbf{s}_a$, an object $\mathbf{s}_o$, and a context image $\mathbf{s}_I$, the generator learns to synthesize a video with background $I$ in which the action $a$ is performed on the object $o$.}
    \label{fig:model}
\end{figure}

Intuitively, for a generated human-object interaction (HOI) video to be realistic, it must: (1) contain the object designated by a semantic label; (2) exhibit the prescribed interaction with that object; (3) be temporally consistent; and (4 -- optional) occur in a specified scene. Based on this intuition, we propose an adversarial learning scheme in which we train a generator network $\mathbf{G}$ with a set of 4 discriminators: (1) a frame discriminator $\mathbf{D}_f$, which encourages the generator to learn spatially coherent visual content; (2) a gradient discriminator $\mathbf{D}_g$, which incentivizes $\mathbf{G}$ to produce temporally consistent frames; (3) a video discriminator $\mathbf{D}_v$, which provides the generator with global spatio-temporal context; and (4) a relational discriminator $\mathbf{D}_r$, which assists the generator in producing correct object layouts in a video. We use pretrained word embeddings~\cite{pennington2014glove} for semantic representations of actions and objects. All discriminators are conditioned on word embeddings of the action ($\mathbf{s}_a$) and object ($\mathbf{s}_o$) and trained simultaneously in an end-to-end manner. Figure~\ref{fig:model} shows an overview of our proposed framework \emph{HOI-GAN}. We now formalize our task and describe each module in detail. 

%
\subsection{Task Formulation} 
Let $\mathbf{s}_a$ and $\mathbf{s}_o$ be word embeddings of an action $a$ and an object $o$, respectively. Furthermore, let $I$ be an image provided as context to the generator. We encode $I$ using an encoder $\mathbf{E}_v$ to obtain a visual embedding $\mathbf{s_\textit{I}}$, which we refer to as a context vector. Our goal is to generate a video $V=(V^{(i)})_{i=1}^T$ of length $T$ depicting the action $a$ performed on the object $o$ with context image $I$ as the background of $V$. To this end, we learn a function $\mathbf{G}: (\mathbf{z},\mathbf{s}_a, \mathbf{s}_o, \mathbf{s_\textit{I}}) \mapsto V$, where $\mathbf{z}$ is a noise vector sampled from a distribution $p_{\mathbf{z}}$, such as a Gaussian distribution. 

\subsection{Model Description}
We describe the elements of our framework below. Overall, the four discriminator networks, \textit{i.e.}, frame discriminator $\mathbf{D}_f$, gradient discriminator $\mathbf{D}_g$, video discriminator $\mathbf{D}_v$, and relational discriminator $\mathbf{D}_r$ are all involved in a zero-sum game with the generator network $\mathbf{G}$. Refer to the supplementary for implementation details.
\\
\noindent
\textbf{Frame Discriminator.}
The frame discriminator network $\mathbf{D}_f$ learns to distinguish between real and generated frames corresponding to the real video $V_{\textrm{real}}$ and generated video $V_{\textrm{gen}} = \mathbf{G}(\mathbf{z},\mathbf{s}_a,\mathbf{s}_o,\mathbf{s}_I)$ respectively. Each frame in $V_{\textrm{gen}}$ and $V_{\textrm{real}}$ is processed independently using a network consisting of stacked \verb+conv2d+ layers, \textit{i.e.}, 2D convolutional layers followed by spectral normalization~\cite{miyato2018spectral} and leaky ReLU layers~\cite{maas2013rectifier} with $a=0.2$. We obtain a tensor of size $N^{(t)}\times w^{(t)}_0 \times h^{(t)}_0$ ($t=1,2,\ldots,T$), where $N^{(t)}$, $w^{(t)}_0$, and $h^{(t)}_0$ are the channel length, width and height of the activation of the last \verb+conv2d+ layer respectively. We concatenate this tensor with spatially replicated copies of $\mathbf{s}_a$ and $\mathbf{s}_o$, which results in a tensor of size $(\textrm{dim}(\mathbf{s}_a)+\textrm{dim}(\mathbf{s}_o)+N^{(t)})\times w^{(t)}_0 \times h^{(t)}_0$. We then apply another \verb+conv2d+ layer to obtain a $N\times w^{(t)}_0 \times h^{(t)}_0$ tensor. We now perform $1\times1$ convolutions followed by $w^{(t)}_0 \times h^{(t)}_0$ convolutions and a sigmoid to obtain a $T$-dimensional vector corresponding to the $T$ frames of the video $V$. The $i$-th element of the output denotes the probability that the frame $V^{(i)}$ is real.
The objective function of the network $\mathbf{D}_f$ is the loss function:
\begin{equation}
\setlength{\jot}{-2pt}
\begin{split}
    L_f = \frac{1}{2T}\sum_{i=1}^{T} & [\log(\mathbf{D}_f^{(i)}(V_{\textrm{real}};\mathbf{s}_a,\mathbf{s}_o)) +
     \log(1 - \mathbf{D}_f^{(i)}(V_{\textrm{gen}};\mathbf{s}_a,\mathbf{s}_o))],
\end{split}
\end{equation}
where $\mathbf{D}_f^{(i)}$ is the $i$-th element of the output of $\mathbf{D}_f$.
\\
\noindent
\textbf{Gradient Discriminator.}
The gradient discriminator network $\mathbf{D}_g$ enforces temporal smoothness by learning to differentiate between the temporal gradient of a real video $V_{\textrm{real}}$ and a generated video $V_{\textrm{gen}}$. We define the temporal gradient $\nabla_{\!t}\,V$ of a video $V$ with $T$ frames $V^{(1)},\ldots,V^{(T)}$ as pixel-wise differences between two consecutive frames of the video. The $i$-th element of $\nabla_t V$ is defined as: 
\begin{equation}
    \left[\nabla_{\!t}\,V\right]_i = V^{(i+1)} - V^{(i)}, \quad i = 1,2, \ldots, (T-1).
\end{equation}
The architecture of the gradient discriminator $\mathbf{D}_g$ is similar to that of the frame discriminator $\mathbf{D}_f$. The output of $\mathbf{D}_g$ is a $(T-1)$-dimensional vector corresponding to the $(T-1)$ values in gradient $\nabla_{\!t}\,V$. The objective function of $\mathbf{D}_g$ is
\begin{equation}
\setlength{\jot}{-2pt}
\begin{split}
     L_g = \frac{1}{2(T-1)}\sum_{i=1}^{T-1} & [\log(\mathbf{D}_g^{(i)}(\nabla_{\!t}\,V_{\textrm{real}};\mathbf{s}_a,\mathbf{s}_o)) +\\
     &\log(1 - \mathbf{D}_g^{(i)}(\nabla_{\!t}\,V_{\textrm{gen}};\mathbf{s}_a,\mathbf{s}_o))],
\end{split}
\end{equation}
where $\mathbf{D}_g^{(i)}$ is the $i$-th element of the output of $\mathbf{D}_g$.
\\
\noindent
\textbf{Video Discriminator.}
The video discriminator network $\mathbf{D}_v$ learns to distinguish between real videos $V_{\textrm{real}}$ and generated videos $V_{\textrm{gen}}$ by comparing their global spatio-temporal contexts.
The architecture consists of stacked \verb+conv3d+ layers, \textit{i.e.}, 3D convolutional layers followed by spectral normalization~\cite{miyato2018spectral} and leaky ReLU layers~\cite{maas2013rectifier} with $a=0.2$. We obtain a $N \times d_0 \times w_0 \times h_0$ tensor, where $N$, $d_0$, $w_0$, and $h_0$ are the channel length, depth, width, and height of the activation of the last \verb+conv3d+ layer respectively. We concatenate this tensor with spatially replicated copies of $\mathbf{s}_a$ and $\mathbf{s}_o$, which results in a tensor of size $(\textrm{dim}(\mathbf{s}_a)+\textrm{dim}(\mathbf{s}_o)+N)\times d_0 \times w_0 \times h_0$, where $\textrm{dim}(\cdot)$ returns the dimensionality of a vector. We then apply another \verb+conv3d+ layer to obtain a $N\times d_0 \times w_0 \times h_0$ tensor. Finally, we apply a $1\times1\times1$ convolution followed by a $d_0\times w_0 \times h_0$ convolution and a sigmoid to obtain the output, which represents the probability that the video $V$ is real.
The objective function of the network $\mathbf{D}_v$ is the following loss function:
\begin{equation}
\begin{split}
    L_v = \frac{1}{2}[&\log(\mathbf{D}_v(V_{\textrm{real}};\mathbf{s}_a,\mathbf{s}_o)) +
    \log(1 - \mathbf{D}_v(V_{\textrm{gen}};\mathbf{s}_a,\mathbf{s}_o))].
\end{split}
\end{equation}

\noindent
\textbf{Relational Discriminator.}
\begin{figure}[t]
    \centering
    \includegraphics[width=0.48\textwidth]{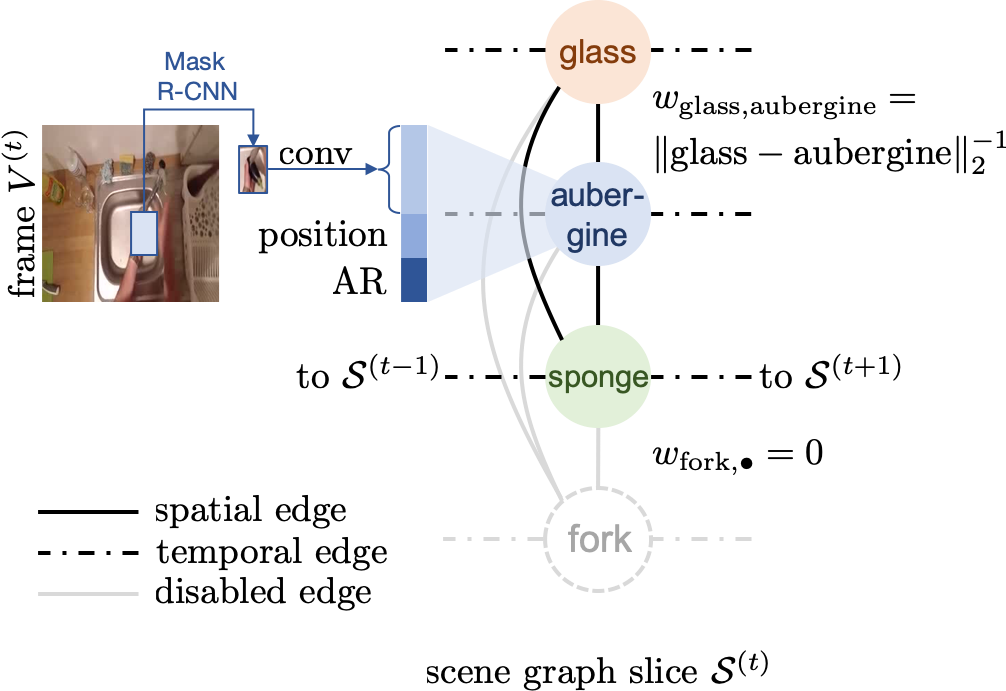}
    \caption{\textbf{Relational Discriminator.} The relational discriminator $\mathbf{D}_r$ leverages a spatio-temporal scene graph to distinguish between object layouts in videos. Each node contains convolutional embedding, position and aspect ratio (AR) of the object crop obtained from MaskRCNN. The nodes are connected in space and time and edges are weighted based on their inverse distance. Edge weights of (dis)appearing objects are 0.
    }
    \label{fig:rel_discriminator}
\end{figure}
In addition to the three pixel-centric discriminators above, we also propose a novel object-centric discriminator $\mathbf{D}_r$. Driven by a spatio-temporal scene graph, this relational discriminator learns to distinguish between scene layouts of real videos $V_{\textrm{real}}$ and generated videos $V_{\textrm{gen}}$ (Figure~\ref{fig:rel_discriminator}). 

Specifically, we build a spatio-temporal scene graph $\mathcal{S} = (\mathcal{N},\mathcal{E})$ from $V$, where the nodes and edges are represented by $\mathcal{N}$ and $\mathcal{E}$ respectively. We assume one node per object per frame. Each node is connected to all other nodes in the same frame, referred to as spatial edges. In addition, to represent temporal evolution of objects, each node is connected to the corresponding nodes in the adjacent frames that also depict the same object, referred to as temporal edges. To obtain the node representations, we crop the objects in $V$ using Mask-RCNN~\cite{he2017mask}, 
compute a convolutional embedding for them, and augment the resulting vectors with the aspect ratio (AR) and position of the corresponding bounding boxes. The weights of spatial edges in $\mathcal{E}$ are given by inverse Euclidean distances between the centers of these bounding boxes corresponding to the object appearing in the frame. The weights of the temporal edges is set to 1 by default. When an object is not present in a frame (but appears in the overall video), spatial edges connecting to the object will be absent by design. This is implemented by setting the weights to 0 depicting distance between the objects as $\infty$. Similarly, if an object does not appear in the adjacent frame, the temporal edge is set to 0. In case of multiple objects of the same category, the correspondence is established based on the location in the adjacent frames using nearest neighbour data association. 

The relational discriminator $\mathbf{D}_r$ operates on this scene graph $\mathcal{S}$ by virtue of a graph convolutional network (GCN)~\cite{kipf2017semi} followed by stacking and average-pooling of the resulting node representations along the time axis. We then concatenate this tensor with spatially replicated copies of $\mathbf{s}_a$ and $\mathbf{s}_o$ to result in a tensor of size $(\textrm{dim}(\mathbf{s}_a)+\textrm{dim}(\mathbf{s}_o)+N^{(t)}) \times w^{(t)}_0 \times h^{(t)}_0$. As before, we then apply convolutions and sigmoid to obtain the final output which denotes the probability of the scene graph belonging to a real video.
The objective function of the network $\mathbf{D}_r$ is given by
\begin{equation}
\begin{split}
     L_r = \frac{1}{2} [& \log(\mathbf{D}_r( \mathcal{S}_{\textrm{real}};\mathbf{s}_a,\mathbf{s}_o)) +
     \log(1 - \mathbf{D}_r(\mathcal{S}_{\textrm{gen}};\mathbf{s}_a,\mathbf{s}_o))] .
\end{split}
\end{equation}

\noindent
\textbf{Generator.} Given the semantic embeddings $\mathbf{s}_a$, $\mathbf{s}_o$ of action and object labels respectively, and context vector $\mathbf{s_\textit{I}}$, the generator network $\mathbf{G}$ learns to generate video $V_{gen}$ consisting of T frames (RGB) of height $H$ and width $W$. We concatenate noise $\mathbf{z}$ with the conditions, namely,  $\mathbf{s}_a$, $\mathbf{s}_o$, and $\mathbf{s}_\textit{I}$.  We provide this concatenated vector as the input to the network $\textbf{G}$. The network comprises stacked \verb+deconv3d+ layers, \textit{i.e.}, 3D transposed convolution layers each followed by Batch Normalization~\cite{ioffe15bn} and leaky ReLU layers~\cite{maas2013rectifier} with $a=0.2$ except the last convolutional layer which is followed by a Batch Normalization layer \cite{ioffe15bn} and a \verb+tanh+ activation layer. The network is optimized according to the following objective function: 
\begin{equation}
\begin{split}
     L_{gan} & = \frac{1}{T}\sum_{i=1}^{T}[\log(1 - \mathbf{D}_f^{(i)}(V_{\textrm{gen}};\mathbf{s}_a,\mathbf{s}_o))] + \\
     &\frac{1}{(T-1)}\sum_{i=1}^{T-1}[\log(1 - \mathbf{D}_g^{(i)}( \nabla_{\!t}\,V_{\textrm{gen}};\mathbf{s}_a,\mathbf{s}_o))] + \\
     & \log(1 - \mathbf{D}_v(V_{\textrm{gen}};\mathbf{s}_a,\mathbf{s}_o)) + 
     \log(1 - \mathbf{D}_r(\mathcal{S}_{\textrm{gen}};\mathbf{s}_a,\mathbf{s}_o)).
\end{split}
\end{equation}


\section{Experiments}
We conduct quantitative and qualitative analysis to demonstrate the effectiveness of the proposed framework HOI-GAN for the task of zero-shot generation of human-object interaction (HOI) videos. 
\subsection{Datasets and Data Splits}
We use two datasets for our experiments: EPIC-Kitchens \cite{Damen2018EPICKITCHENS} and 20BN-Something-Something V2 \cite{goyal2017something}. Both of these datasets comprise a diverse set of HOI videos ranging from simple translational motion of objects (\textit{e.g.}\ push, move) and rotation (\textit{e.g.}\ open) to transformations in state of objects (\textit{e.g.}\ cut, fold). 
Therefore, these datasets, with their wide ranging variety and complexity, provide a challenging setup for evaluating HOI video generation models.

\textbf{EPIC-Kitchens} \cite{Damen2018EPICKITCHENS} contains egocentric videos of activities in several kitchens. A video clip $V$ is annotated with action label $a$ and object label $o$ (\textit{e.g.} open microwave, cut apple, move pan) along with a set of bounding boxes $\mathcal{B}$ (one per frame) for objects that the human interacts with while performing the action. There are around 40k instances in the form of $(V, a, o, \mathcal{B})$ across 352 objects and 125 actions. We refer to this dataset as EPIC hereafter. 

\textbf{20BN-Something-Something V2} \cite{goyal2017something} contains videos of daily activities performed by humans. A video clip $V$ is annotated with a label $l$, an action template and object(s) on which the action is applied (\textit{e.g.}\ `hitting ball with racket' has action template `hitting something with something'). There are 220,847 training instances of the form $(V, l)$ spanning 30,408 objects and 174 action templates. To transform $l$ to action-object label pair $(a,o)$, we use NLTK POS-tagger. 
We consider the verb tag (after stemming) in $l$ as action label $a$. We observe that all instances of $l$ begin with the present continuous form of $a$ which is acting upon the subsequent noun. Therefore, we use the noun that appears immediately after the verb as object ${o}$. Hereafter, we refer to the transformed dataset in the form of $(V,a,o)$ as SS. 
    
\noindent \textbf{Splitting by Compositions.}
We believe it is reasonable to only generate combinations that are semantically feasible, and do so by only using action-object pairs seen in the original datasets. We use a subset of action-object pairs as testing pairs -- these pairs are not seen during training but are present in the original dataset, hence are semantically feasible. To make the dataset training / testing splits suitable for our zero-shot compositional setting, we first merge the data samples present in the default train and validation sets of the dataset. 
We then split the combined dataset into training set and test set based on the condition that all the unique object and action labels in appear in the training set, however, any composition of action and object present in the test set is absent in training set and vice versa.
We provide further details of the splits for both datasets EPIC and SS in the supplementary.

\setlength{\tabcolsep}{4.7pt}
\renewcommand{\arraystretch}{1.0}
\begin{table}[!t]
\centering
\caption{\textbf{Generation Scenarios.} Description of the conditional inputs for the two generation scenarios GS1 \& GS2 used for evaluation. \cmark~denotes `Yes', \xmark~denotes `No'. 
}
\scalebox{1.0}{
\begin{tabular}{l@{\hskip 8mm}c@{\hskip 7mm}c}
\toprule
\centering
\textbf{Target Conditions} & \textbf{GS1} & \textbf{GS2}\\
\midrule
Target action $a$ seen during training & \cmark  & \cmark
\\
Target object $o$ seen during training & \cmark  & \cmark
\\
Background of target context $I$ seen during training & \xmark  & \cmark
\\
Object mask in target context $I$ corresponds to target object $o$ & \cmark & \xmark
\\
Target action $a$ seen with target context $I$ during training & \xmark & \cmark / \xmark 
\\
Target object $o$ seen with target context $I$ during training & \xmark & \xmark
\\
Target action-object composition ($a$-$o$) seen during training & \xmark & \xmark
\\
\bottomrule
\end{tabular}
}
\label{tab:scenario}
\end{table}

\noindent \textbf{Generation Scenarios.} Recall that the generator network in the HOI-GAN framework (Fig.~\ref{fig:model}) has 3 conditional inputs, namely, action embedding, object embedding, and context frame $I$. The context frame serves as the background in the scene. Thus, to provide this context frame during training, we apply a binary mask $M^{(1)}$ corresponding to the first frame $V^{(1)}$ of a real video as $I = (\mathbbm{1}-M^{(1)}) \odot V^{(1)}$, where $\mathbbm{1}$ represents a matrix of size $M^{(1)}$ containing all ones and $\odot$ denotes elementwise multiplication. This mask $M^{(1)}$ contains ones in regions (either rectangular bounding boxes or segmentation masks) corresponding to the objects (non-\textit{person} classes) detected using MaskRCNN\cite{he2017mask} and zeros for other regions. 
Intuitively, this helps ensure the generator learns to map the action and object embeddings to relevant visual content in the HOI video.

During testing, to evaluate the generator's capability to synthesize the right human-object interactions, we provide a background frame as described above. 
This background frame can be selected from either the test set or training set, and can be suitable or unsuitable for the target action-object composition. 
To capture these possibilities, we design two different generation scenarios.
Specifically, in \textit{Generation Scenario 1 (GS1)}, the input context frame $I$ is the masked first frame of a video from the test set corresponding to the target action-object composition (unseen during training). In \textit{Generation Scenario 2 (GS2)}, $I$ is the masked first frame of a video from the training set which depicts an object other than the target object. The original action in this video could be same or different than the target action. See Table~\ref{tab:scenario} for the contrast between the scenarios. 

As such, in GS1, the generator receives a context that it has not seen during training but the context (including object mask) is consistent with the target action-object composition it is being asked to generate. 
In contrast, in GS2, the  generator receives a context frame that it has seen during training but is not consistent with the action-object composition it is being asked to generate. Particularly, the object mask in the context does not correspond to the target object. Although the background is seen, the model has to evolve the background in ways different from training samples to make it suitable for the target composition.
Thus, these generation scenarios help illustrate that the generator indeed generalizes over compositions.

\subsection{Evaluation Setup}
Evaluation of image/video quality is inherently challenging, thus, we use both quantitative and qualitative metrics.
\setlength{\tabcolsep}{1.5pt}
\renewcommand{\arraystretch}{0}
\begin{figure*}[t!]
\centering
\begin{tabular}{@{\hskip 2mm}lp{1.3cm} @{\hskip 1mm}c @{\hskip 3mm} c}
\toprule
Scenario & $(a,o)$ & Context & Generated Output\\
\cmidrule(lr){1-3}\cmidrule(lr){4-4}
\begin{tabular}{l}
GS1
\end{tabular}
&
\begin{tabular}{p{1.2cm}}
take spoon (EPIC)
\end{tabular}
&
\begin{tabular}{c}
\includegraphics[scale=0.8]{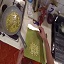}
\end{tabular}
&
\begin{tabular}{cccccc}
\includegraphics[scale=0.8]{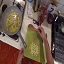}
&
\includegraphics[scale=0.8]{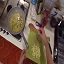}
&
\includegraphics[scale=0.8]{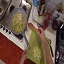}
&
\includegraphics[scale=0.8]{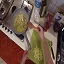}
&
\includegraphics[scale=0.8]{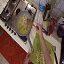}
\end{tabular}
\\
\midrule
\begin{tabular}{l}
GS1
\end{tabular}
&
\begin{tabular}{p{1.0cm}}
hold cup (SS)
\end{tabular}
&
\begin{tabular}{c}
\includegraphics[scale=0.8]{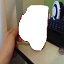}
\end{tabular}
&
\begin{tabular}{cccccc}
\includegraphics[scale=0.8]{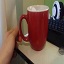}
&
\includegraphics[scale=0.8]{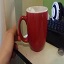}
&
\includegraphics[scale=0.8]{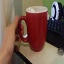}
&
\includegraphics[scale=0.8]{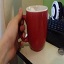}
&
\includegraphics[scale=0.8]{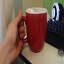}
\end{tabular}
\\
\midrule
\begin{tabular}{l}
GS2
\end{tabular}
&
\begin{tabular}{p{1.2cm}}
move broccoli (EPIC)
\end{tabular}
&
\begin{tabular}{c}
\includegraphics[scale=0.8]{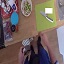}
\end{tabular}
&
\begin{tabular}{cccccc}
\includegraphics[scale=0.8]{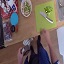}
&
\includegraphics[scale=0.8]{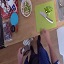}
&
\includegraphics[scale=0.8]{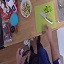}
&
\includegraphics[scale=0.8]{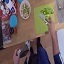}
&
\includegraphics[scale=0.8]{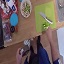}
\end{tabular}
\\
\midrule
\begin{tabular}{l}
GS2
\end{tabular}
&
\begin{tabular}{p{1.2cm}}
put \mbox{apple} (SS)
\end{tabular}
&
\begin{tabular}{c}
\includegraphics[scale=0.8]{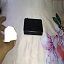}
\end{tabular}
&
\begin{tabular}{cccccc}
\includegraphics[scale=0.8]{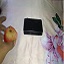}
&
\includegraphics[scale=0.8]{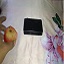}
&
\includegraphics[scale=0.8]{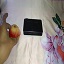}
&
\includegraphics[scale=0.8]{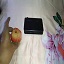}
&
\includegraphics[scale=0.8]{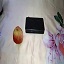}
\end{tabular}
\\
\bottomrule
\end{tabular}
\caption{\textbf{Qualitative Results}: Videos generated using our best version of HOI-GAN using embeddings for action $(a)$-object $(o)$ composition and the context frame. We show 5 frames of the video clip generated for both generation scenarios GS1 and GS2. The context frame in GS1 is obtained from a video in the test set depicting an action-object composition same as the target one. 
The context frame for GS2 scenarios shown here are from videos depicting \emph{``take carrot"} (for row 3) and \emph{``put bowl"} (for row 4). Best viewed in color on desktop. Refer to supplementary section for additional videos generated using HOI-GAN.}
\label{fig:qual}
\end{figure*}

\noindent
\textbf{Quantitative Metrics.} Inception Score \textbf{(I-score)} \cite{salimans2016improved} is a widely used metric for evaluating image generation models. For images $x$ with labels $y$, I-score is defined as $\exp(\mathbf{KL}(\rho(y|x)||\rho(y)))$ where $\rho(y|x)$ is the conditional label distribution of an ImageNet\cite{ILSVRC15} -pretrained Inception model \cite{szegedy2016rethinking}. 
We adopted this metric for video quality evaluation. We fine-tune a Kinetics \cite{carreira2017quo}-pretrained video classifier ResNeXt-101 \cite{xie2017aggregated} for each of our source datasets and use it for calculating I-score (higher is better). It is based on one of the state-of-the-art video classification architectures. We used the same evaluation setup for the baselines and our model to ensure a fair comparison. 

In addition, we believe that measuring realism explicitly is more relevant for our task as the generation process can be conditioned on any context frame arbitrarily to obtain diverse samples. Therefore, in addition to \textit{I-score}, we also analyze the first and second terms of the KL divergence separately. We refer to these terms as: (1) Saliency score or \textbf{S-score} (lower is better) to specifically measure realism, and (2) Diversity score or \textbf{D-score} (higher is better) to indicate the diversity in generated samples. A smaller value of S-score implies that the generated videos are more realistic as the classifier is very confident in classifying the generated videos.  Specifically, the saliency score will have a low value (low is good) only when the classifier is confidently able to classify the generated videos into action-object categories matching the conditional input composition (action-object), thus indicating realistic instances of the required target interaction. In fact, even if a model generates realistic-looking videos but depicts an action-object composition not corresponding to the conditional action-object input, the saliency score will have high values. Finally, a larger value of D-score implies the model generates diverse samples.


\noindent
\textbf{Human Preference Score.} 
We conduct a user study for evaluating the quality of generated videos. In each test, we present the participants with two videos generated by two different algorithms and ask which among the two better depicts the given activity, \textit{i.e.}, action-object composition (\textit{e.g.}\ lift fork). We evaluate the performance of an algorithm as the overall percentage of tests in which that algorithm's outputs are preferred. This is an aggregate measure over all the test instances across all participants.

\noindent \textbf{Baselines.}
We compare HOI-GAN with three state-of-the-art video generation approaches: (1) VGAN \cite{vondrick2016generating}, (2) TGAN, \cite{saito2017temporal} and (3) MoCoGAN \cite{tulyakov2018mocogan}. We develop the conditional variants of VGAN and TGAN from the descriptions provided in their papers. We refer to the conditional variants as C-VGAN and C-TGAN respectively. We observed that these two models saturated easily in the initial iterations, thus, we added dropout in the last layer of the discriminator network in both models. MoCoGAN focuses on disentangling motion and content in the latent space and is the closest baseline. We use the code provided by the authors. 

\subsection{Results}\label{sec:results} 
\noindent
Next, we discuss the results of our qualitative and quantitative evaluation. 

\setlength{\tabcolsep}{4pt}
\renewcommand{\arraystretch}{1}
\begin{table*}[t]
\centering
\caption{\textbf{Quantitative Evaluation.} Comparison of HOI-GAN with C-VGAN, C-TGAN, and MoCoGAN baselines. We distinguish training of HOI-GAN with bounding boxes (\textit{bboxes}) and segmentation masks (\emph{masks}). Arrows indicate whether lower ($\downarrow$) or higher ($\uparrow$) is better. [I: inception score; S: saliency score; D: diversity score]}
\scalebox{0.9}{
\begin{tabular}{@{\hskip 1mm}l@{\hskip 0.5mm}l@{\hskip 0.1mm}c@{\hskip 1mm}c@{\hskip 1mm}c@{\hskip1mm}c}
\toprule
&\multirow{3}{*}{Model} & \multicolumn{2}{c}{EPIC} & \multicolumn{2}{c}{SS} \\
\cmidrule(lr){3-4}\cmidrule(lr){5-6}
& & GS1 & GS2 & GS1 & GS2 \\
\cmidrule(lr){3-3}\cmidrule(lr){4-4}\cmidrule(lr){5-5}\cmidrule(lr){6-6}
& & 
I$\uparrow$ {\hskip 1mm} S$\downarrow$ {\hskip 1.2mm} D$\uparrow$
& 
I$\uparrow$ {\hskip 1mm} S$\downarrow$ {\hskip 1.2mm} D$\uparrow$
&
I$\uparrow$ {\hskip 1mm} S$\downarrow$ {\hskip 1.2mm} D$\uparrow$
& 
I$\uparrow$ {\hskip 1mm} S$\downarrow$ {\hskip 1.2mm} D$\uparrow$
\\
\midrule
& \small{C-VGAN}~\cite{vondrick2016generating}
&
\begin{tabular}{c c c}
1.8 & 30.9 & 0.2\\ 
\end{tabular}
&
\begin{tabular}{c c c}
1.4 & 44.9& 0.3\\ 
\end{tabular}
&
\begin{tabular}{c c c}
2.1 & 25.4 & 0.4\\ 
\end{tabular}
&
\begin{tabular}{c c c}
1.8 & 40.5 & 0.3\\ 
\end{tabular}
\\
& \small{C-TGAN}~\cite{saito2017temporal}
&
\begin{tabular}{c c c}
2.0 & 30.4 & 0.6\\ 
\end{tabular}
&
\begin{tabular}{c c c}
1.5 & 35.9 & 0.4\\ 
\end{tabular}
&
\begin{tabular}{c c c}
2.2 & 28.9 & 0.6\\ 
\end{tabular}
&
\begin{tabular}{c c c}
1.6 & 39.7 & 0.5\\ 
\end{tabular}
\\
& \small{MoCoGAN}~\cite{tulyakov2018mocogan}
&
\begin{tabular}{c c c}
2.4 & 30.7 & 0.5\\ 
\end{tabular}
&
\begin{tabular}{c c c}
2.2 & 31.4 & 1.2
\end{tabular}
&
\begin{tabular}{c c c}
2.8 & 17.5 & 1.0\\ 
\end{tabular}
&
\begin{tabular}{c c c}
2.4 & 33.7 & 1.4
\end{tabular}
\\
\midrule
\multirow{2}{*}{\rotatebox[origin=c]{90}{{\footnotesize (ours)}}}
& \small{HOI-GAN (bboxes)} &
\begin{tabular}{c c c}
6.0 & 14.0 & 3.4
\\ 
\end{tabular}
&
\begin{tabular}{c c c}
5.7 & 20.8 & 4.0\\ 
\end{tabular}
&
\begin{tabular}{c c c}
6.6 & 12.7 & 3.5 \\
\end{tabular}
&
\begin{tabular}{c c c}
6.0& 15.2&2.9\\
\end{tabular}
\\
& \small{HOI-GAN (masks)} &
\begin{tabular}{c c c}
{\hskip 0.8mm}\bf 6.2 & {\hskip -1mm}\bf 13.2 & {\hskip -1mm}\bf 3.7\\
\end{tabular}
&
\begin{tabular}{c c c}
{\hskip 0.8mm}\textbf{5.2} & 
{\hskip -0.8mm}\textbf{18.3} & 
{\hskip -0.8mm}\textbf{3.5}\\
\end{tabular}
&
\begin{tabular}{c c c}
{\hskip 0.8mm}\textbf{8.6} & 
{\hskip -0.8mm}\textbf{11.4} & 
{\hskip -0.8mm}\textbf{4.4}\\
\end{tabular}
&
\begin{tabular}{c c c}
{\hskip 0.8mm}\textbf{7.1} & 
{\hskip -0.8mm}\textbf{14.7} & 
{\hskip -0.8mm}\textbf{4.0}\\
\end{tabular}
\\
\bottomrule
\end{tabular}}
\label{tab:combined}
\end{table*}

\setlength{\tabcolsep}{4.5pt}
\renewcommand{\arraystretch}{1.0}
\begin{table*}[t]
\caption{\textbf{Ablation Study.} We evaluate the contributions of our pixel-centric losses (F,G,V) and relational losses (first block vs. second block) by conducting ablation study on HOI-GAN (masks). The last row corresponds to the overall proposed model.[F: frame discriminator $\mathbf{D}_f$; G: gradient discriminator $\mathbf{D}_g$; V: video discriminator $\mathbf{D}_v$; R: relational discriminator $\mathbf{D}_r$]}
\centering
\scalebox{0.9}{
\begin{tabular}{l@{\hskip 1mm}l@{\hskip 0mm}c@{\hskip 1mm}c@{\hskip 1mm}c@{\hskip 1mm}c}
\toprule
& \multirow{3}{*}{Model} & \multicolumn{2}{c}{EPIC} & \multicolumn{2}{c}{SS} \\
\cmidrule(lr){3-4}\cmidrule(lr){5-6}
& & GS1 & GS2 & GS1 & GS2 \\
\cmidrule(lr){3-3}\cmidrule(lr){4-4}\cmidrule(lr){5-5}\cmidrule(lr){6-6}
& & 
I$\uparrow$ {\hskip 1.4mm} S$\downarrow$ {\hskip 1.4mm} D$\uparrow$
& 
I$\uparrow$ {\hskip 1.4mm} S$\downarrow$ {\hskip 1.4mm} D$\uparrow$
&
I$\uparrow$ {\hskip 1.4mm} S$\downarrow$ {\hskip 1.4mm} D$\uparrow$
& 
I$\uparrow$ {\hskip 1.4mm} S$\downarrow$ {\hskip 1.4mm} D$\uparrow$
\\
\midrule
\multirow{3}{*}{\rotatebox[origin=c]{90}{{$-$\small{R}}}}
& \small {HOI-GAN (F)} &
\begin{tabular}{c c c}
1.4 & 44.2 & 0.2\\
\end{tabular}
&
\begin{tabular}{c c c}
1.1 & 47.2 & 0.3\\
\end{tabular}
&
\begin{tabular}{c c c}
1.8 & 34.7 & 0.4\\
\end{tabular}
&
\begin{tabular}{c c c}
1.5 & 39.5 & 0.3\\
\end{tabular}
\\
& \small {HOI-GAN (F$+$G)}
&
\begin{tabular}{c c c}
2.3 & 25.6 & 0.7\\
\end{tabular}
&
\begin{tabular}{c c c}
1.9 & 30.7 & 0.5\\
\end{tabular}
&
\begin{tabular}{c c c}
3.0 & 24.5 & 0.9\\
\end{tabular}
&
\begin{tabular}{c c c}
2.7 & 28.8 & 0.7\\
\end{tabular}
\\
& \small {HOI-GAN (F$+$G$+$V)}
&
\begin{tabular}{c c c}
2.8 & 21.2 & 1.3\\
\end{tabular}
&
\begin{tabular}{c c c}
2.6 & 29.7 & 1.7\\
\end{tabular}
&
\begin{tabular}{c c c}
3.3 & 18.6 & 1.2\\
\end{tabular}
&
\begin{tabular}{c c c}
3.0 & 20.7 & 1.0\\
\end{tabular}
\\
\midrule
\multirow{3}{*}{\rotatebox[origin=c]{90}{{$+$\small{R}}}}
& \small {HOI-GAN (F)}
&
\begin{tabular}{c c c}
2.4 & 24.9 & 0.8\\
\end{tabular}
&
\begin{tabular}{c c c}
2.2 & 26.0 & 0.7\\
\end{tabular}
&
\begin{tabular}{c c c}
3.1 & 20.3 & 1.0\\
\end{tabular}
&
\begin{tabular}{c c c}
2.9 & 27.7 & 0.9\\
\end{tabular}
\\
& \small {HOI-GAN (F$+$G)}
&
\begin{tabular}{c c c}
5.9 & 15.4 &3.5\\
\end{tabular}
&
\begin{tabular}{c c c}
4.8 & 21.3 & 3.3\\
\end{tabular}
&
\begin{tabular}{c c c}
7.4 & 12.1 & 3.5\\
\end{tabular}
&
\begin{tabular}{c c c}
5.4 & 19.2  & 3.4\\
\end{tabular}
\\
& \small {HOI-GAN (F$+$G$+$V)}
&
\begin{tabular}{c c c}
6.2 & 13.2 & 3.7\\
\end{tabular}
&
\begin{tabular}{c c c}
5.2 & 18.3 & 3.5\\
\end{tabular}
&
\begin{tabular}{c c c}
8.6 & 11.4 & 4.4\\
\end{tabular}
&
\begin{tabular}{c c c}
7.1 & 14.7 & 4.0\\
\end{tabular}
\\
\bottomrule
\end{tabular}}
\label{tab:ablation}
\end{table*}
\renewcommand{\arraystretch}{1.0}

\noindent
\textbf{Comparison with Baselines.} 
As shown in Table~\ref{tab:combined}, HOI-GAN with different conditional inputs outperforms C-VGAN and C-TGAN by a wide margin in both generation scenarios. In addition, our overall model shows considerable improvement over MoCoGAN, while MoCoGAN has comparable scores to some ablated versions of our models (where gradient discriminator and/or relational discriminator is missing). Furthermore, we varied the richness of the masks in the conditional input context frame ranging from bounding boxes to segmentation masks obtained corresponding to non-\emph{person} classes using MaskRCNN framework \cite{he2017mask}. 
We observe that providing masks during training leads to slight improvements in both scenarios as compared to using bounding boxes (refer to Table~\ref{tab:combined}). We also show the samples generated using the best version of HOI-GAN for the two generation scenarios (Figure~\ref{fig:qual}). See supplementary for more generated samples and detailed qualitative analysis.

\noindent
\textbf{Ablation Study.} To illustrate the impact of each discriminator in generating HOI videos, we conduct ablation experiments (refer to Table~\ref{tab:ablation}). We observe that the addition of temporal information using the gradient discriminator and spatio-temporal information using the video discriminator lead to improvement in generation quality. In particular, the addition of our scene graph based relational discriminator leads to considerable improvement in generation quality resulting in more realistic videos (refer to second block in Table~\ref{tab:ablation}). 
Additional quantitative studies and results are in the supplementary.

\noindent
\textbf{Human Evaluation.}
We recruited 15 sequestered participants for our user study. We randomly chose 50 unique categories and chose generated videos for half of them from generation scenario GS1 and the other half from GS2. For each category, we provided three instances, each containing a pair of videos; one generated using a baseline model and the other using HOI-GAN. For each instance, at least 3 participants (ensuring inter-rater reliability) are asked to choose the video that best depicts the given category. 
The (aggregate) human preference scores for our model versus the baselines range between 69-84\% for both generation scenarios (refer Table~\ref{tab:human}). These results indicate that HOI-GAN generates more realistic videos than the baselines.
\setlength{\tabcolsep}{10pt}
\renewcommand{\arraystretch}{0.8}
\begin{table}[!t]
\centering
\caption{\textbf{Human Evaluation.} Human Preference Score (\%) for scenarios GS1 and GS2. All the results have p-value less than 0.05 implying statistical significance.}
\begin{tabular}{lcc}
\toprule
Ours / Baseline & GS1 & GS2\\
\midrule
\small{HOI-GAN / MoCoGAN} & \textbf{71.7}/28.3 &\textbf{69.2}/30.8 \\
\small{HOI-GAN / C-TGAN} & \textbf{75.4}/34.9& \textbf{79.3}/30.7\\
\small{HOI-GAN / C-VGAN} & \textbf{83.6}/16.4& \textbf{80.4}/19.6\\
\bottomrule
\end{tabular}
\label{tab:human}
\end{table}
\setlength{\tabcolsep}{2pt}
\renewcommand{\arraystretch}{0}
\begin{figure}[!t]
\centering
\begin{tabular}{p{1.5cm} @{\hskip 1mm}c c}
\toprule
$(a,o)$ & Context & Generated Output\\
\cmidrule(lr){1-2}\cmidrule(lr){3-3}
\begin{tabular}{p{1.2cm}}
open micro-wave
\end{tabular}
&
\begin{tabular}{c}
\includegraphics[scale=0.4]{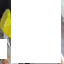}
\end{tabular}
&
\begin{tabular}{cccc}
\includegraphics[scale=0.8]{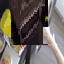}
&
\includegraphics[scale=0.8]{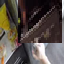}
&
\includegraphics[scale=0.8]{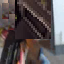}
&
\includegraphics[scale=0.8]{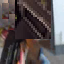}
\end{tabular}
\\
\midrule
\begin{tabular}{p{1.2cm}}
cut peach
\end{tabular}
&
\begin{tabular}{c}
\includegraphics[scale=0.8]{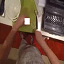}
\end{tabular}
&
\begin{tabular}{cccc}
\includegraphics[scale=0.8]{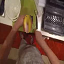}
&
\includegraphics[scale=0.8]{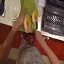}
&
\includegraphics[scale=0.8]{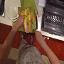}
&
\includegraphics[scale=0.8]{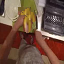}
\end{tabular}
\\
\bottomrule
\end{tabular}
\caption{\textbf{Failure Cases.} Videos generated using HOI-GAN corresponding to the given action-object composition $(a,o)$ and the context frame. We show 4 frames of the videos.}
\label{fig:failure_sample}
\end{figure}

\noindent
\textbf{Failure Cases.}
We discuss the limitations of our framework using qualitative examples shown in Figure~\ref{fig:failure_sample}. For \emph{``open microwave"}, we observe that although HOI-GAN is able to generate conventional colors for a microwave, it shows limited capability to hallucinate such large objects. For \emph{``cut peach"} (Figure~\ref{fig:failure_sample}), 
the generated sample shows that our model can learn the increase in count of partial objects corresponding to the action cut and yellow-green color of a peach. However, as the model has not observed the interior of a peach during training (as \emph{cut peach} was not in training set), it is unable to create realistic transformations in the state of \emph{peach} that show the interior clearly. We provide additional discussion on the failure cases in the supplementary. 


%

\section{Conclusion}
In this paper, we introduced the task of zero-shot HOI video generation, \textit{i.e.}, generating human-object interaction (HOI) videos corresponding to unseen action-object compositions, having seen the target action and target object independently. Towards this goal, we proposed the HOI-GAN framework that uses a novel multi-adversarial learning scheme and demonstrated its effectiveness on challenging HOI datasets. We show that an object-level relational discriminator is an effective means for GAN-based generation of interaction videos. Future work can benefit from our idea of using relational adversaries to synthesize more realistic videos. We believe relational adversaries to be relevant beyond video generation in tasks such as layout-to-image translation.
\\

\noindent
\textbf{Acknowledgements.} This work was done when Megha Nawhal was an intern at Borealis AI. We would like to thank the Borealis AI team for participating in our user study. 

%
%
\bibliographystyle{splncs04}
\bibliography{paper_review}
\clearpage
\section{Supplementary}
This section contains the supplementary information supporting the content in the main paper. 
\begin{itemize} 
\setlength\itemsep{-0.02in}
\item Qualitative evaluation and analysis of HOI-GAN to supplement Section 4.3.
\item Qualitative evaluation of baselines: samples generated using baselines to supplement Section 4.3.
\item Additional quantitative evaluation of our model to supplement Section 4.3.
\item Details of preprocessing and data splits for each dataset to supplement Section 4.1.
\item Implementation details of our model to supplement Section 3.
\end{itemize}

\subsection{Qualitative Evaluation and Analysis of HOI-GAN}
Please view the samples together in the video on this webpage \url{http://www.sfu.ca/~mnawhal/projects/zs_hoi_generation.html}.

\renewcommand{\arraystretch}{1}
\setlength{\tabcolsep}{4pt}
\begin{figure}[!h]
\begin{tabular}{m{1cm}c @{\hskip 0.01in}c @{\hskip 0.01in}c}
\toprule

\hspace{2mm}$(a,o)$ & Context & Generated Output & Original (from test set)\\
\cmidrule(lr){1-1}\cmidrule(lr){2-2}\cmidrule(lr){3-3}\cmidrule(lr){4-4}

\begin{tabular}{m{1cm}} take spoon \end{tabular} &
\begin{tabular}{c} \includegraphics[scale=0.6]{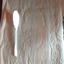}
\end{tabular}
&

\begin{tabular}{c @{\hspace{0.5mm}} c @{\hspace{0.5mm}}c}
\includegraphics[scale=0.6]{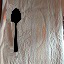}
&
\includegraphics[scale=0.6]{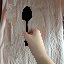}
&
\includegraphics[scale=0.6]{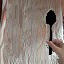}
\\
\end{tabular}

&

\begin{tabular}{c @{\hspace{0.5mm}} c @{\hspace{0.5mm}} c}
\includegraphics[scale=0.6]{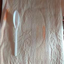}
&
\includegraphics[scale=0.6]{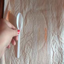}
&
\includegraphics[scale=0.6]{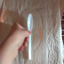}
\\
\end{tabular}
\\
\midrule

\begin{tabular}{m{1cm}}move book\end{tabular} &
\begin{tabular}{c}\includegraphics[scale=0.6]{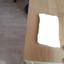}\end{tabular}
&

\begin{tabular}{c @{\hspace{0.5mm}} c @{\hspace{0.5mm}} c }
\includegraphics[scale=0.6]{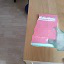}
&
\includegraphics[scale=0.6]{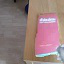}
&
\includegraphics[scale=0.6]{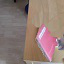}
\\
\end{tabular}

&

\begin{tabular}{c @{\hspace{0.5mm}} c @{\hspace{0.5mm}} c }
\includegraphics[scale=0.6]{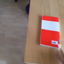}
&
\includegraphics[scale=0.6]{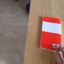}
&
\includegraphics[scale=0.6]{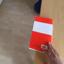}
\\
\end{tabular}
\\
\midrule

\begin{tabular}{m{1cm}}lift toothbrush\end{tabular} &
\begin{tabular}{c}\includegraphics[scale=0.6]{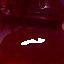}\end{tabular}
&

\begin{tabular}{c @{\hspace{0.5mm}} c @{\hspace{0.5mm}} c }
\includegraphics[scale=0.6]{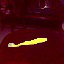}
&
\includegraphics[scale=0.6]{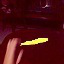}
&
\includegraphics[scale=0.6]{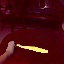}
\\
\end{tabular}

&

\begin{tabular}{c @{\hspace{0.5mm}} c @{\hspace{0.5mm}} c }
\includegraphics[scale=0.6]{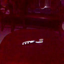}
&
\includegraphics[scale=0.6]{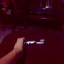}
&
\includegraphics[scale=0.6]{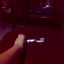}
\\
\end{tabular}
\\
\midrule

\begin{tabular}{m{1.1cm}} put\\ banana\end{tabular} &
\begin{tabular}{c}\includegraphics[scale=0.6]{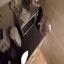}\end{tabular}
&

\begin{tabular}{c @{\hspace{0.5mm}} c @{\hspace{0.5mm}} c }
\includegraphics[scale=0.6]{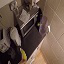}
&
\includegraphics[scale=0.6]{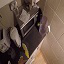}
&
\includegraphics[scale=0.6]{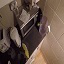}
\\
\end{tabular}

&

\begin{tabular}{c @{\hspace{0.5mm}} c @{\hspace{0.5mm}} c }
\includegraphics[scale=0.6]{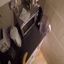}
&
\includegraphics[scale=0.6]{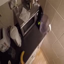}
&
\includegraphics[scale=0.6]{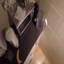}
\\
\end{tabular}
\\
\midrule

\begin{tabular}{m{1cm}}remove cup\end{tabular} &
\begin{tabular}{c}\includegraphics[scale=0.6]{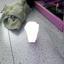}\end{tabular}
&

\begin{tabular}{c @{\hspace{0.5mm}} c @{\hspace{0.5mm}} c }
\includegraphics[scale=0.6]{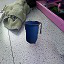}
&
\includegraphics[scale=0.6]{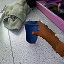}
&
\includegraphics[scale=0.6]{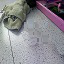}
\\
\end{tabular}

&

\begin{tabular}{c @{\hspace{0.5mm}} c @{\hspace{0.5mm}} c }
\includegraphics[scale=0.6]{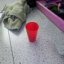}
&
\includegraphics[scale=0.6]{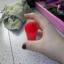}
&
\includegraphics[scale=0.6]{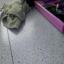}
\\
\end{tabular}
\\
\bottomrule
\end{tabular}
\caption{\textbf{Qualitative Evaluation (GS1).} Samples generated using our model in Generation Scenario 1, \textit{i.e.}, both the target context image and the target action-object $(a,o)$ composition are unseen during training. We provide 3 frames of the generated output and 3 frames of the original video (same context, action, object) from the test set for comparison. }
\label{fig:gs1_samples} 
\end{figure}

\renewcommand{\arraystretch}{1}
\setlength{\tabcolsep}{7pt}
\begin{figure}[!h]
\begin{tabular}{p{3cm} @{\hskip 0.02in} c @{\hskip 0.02in} c}
\toprule
Action-object labels & Context & Generated output\\
\cmidrule(lr){1-1}\cmidrule(lr){2-2}\cmidrule(lr){3-3}
\begin{tabular}{l}
G: lift apple
\\
O: hold banana
\end{tabular}
&
\begin{tabular}{c}
\includegraphics[scale=0.6]{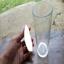}
\end{tabular}
&

\begin{tabular}{c @{\hspace{0.5mm}} c @{\hspace{0.5mm}} c @{\hspace{0.5mm}} c @{\hspace{0.5mm}} c}
\includegraphics[scale=0.6]{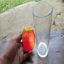}
&
\includegraphics[scale=0.6]{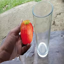}
&
\includegraphics[scale=0.6]{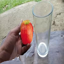}
&
\includegraphics[scale=0.6]{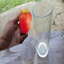}
&
\includegraphics[scale=0.6]{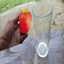}
\\
\end{tabular}
\\

\midrule

\begin{tabular}{l}
G: push scissors
\\
O: pull spoon
\end{tabular}
&

\begin{tabular}{c}
\includegraphics[scale=0.6]{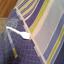}
\end{tabular}
&

\begin{tabular}{c @{\hspace{0.5mm}} c @{\hspace{0.5mm}} c @{\hspace{0.5mm}} c @{\hspace{0.5mm}} c}
\includegraphics[scale=0.6]{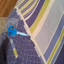}
&
\includegraphics[scale=0.6]{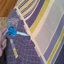}
&
\includegraphics[scale=0.6]{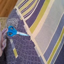}
&
\includegraphics[scale=0.6]{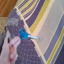}
&
\includegraphics[scale=0.6]{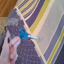}
\\
\end{tabular}
\\

\midrule

\begin{tabular}{l}
G: cut carrot
\\
O: cut celery
\end{tabular}
&

\begin{tabular}{c}
\includegraphics[scale=0.6]{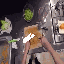}
\end{tabular}
&

\begin{tabular}{c @{\hspace{0.5mm}} c @{\hspace{0.5mm}} c @{\hspace{0.5mm}} c @{\hspace{0.5mm}} c}
\includegraphics[scale=0.6]{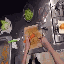}
&
\includegraphics[scale=0.6]{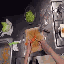}
&
\includegraphics[scale=0.6]{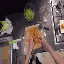}
&
\includegraphics[scale=0.6]{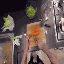}
&
\includegraphics[scale=0.6]{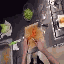}
\\
\end{tabular}
\\
\midrule

\begin{tabular}{l}
G: turn vase
\\
O: move bottle
\end{tabular}
&
\begin{tabular}{c}
\includegraphics[scale=0.6]{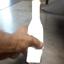}
\end{tabular}
&

\begin{tabular}{c @{\hspace{0.5mm}} c @{\hspace{0.5mm}} c @{\hspace{0.5mm}} c @{\hspace{0.5mm}} c}
\includegraphics[scale=0.6]{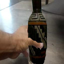}
&
\includegraphics[scale=0.6]{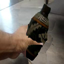}
&
\includegraphics[scale=0.6]{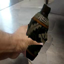}
&
\includegraphics[scale=0.6]{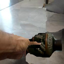}
&
\includegraphics[scale=0.6]{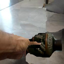}
\\
\end{tabular}
\\

\midrule

\begin{tabular}{l}
G: spin bottle
\\
O: spin remote
\end{tabular}
&
\begin{tabular}{l}
\includegraphics[scale=0.6]{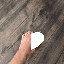}
\end{tabular}
&

\begin{tabular}{c @{\hspace{0.5mm}} c @{\hspace{0.5mm}} c @{\hspace{0.5mm}} c @{\hspace{0.5mm}} c}
\includegraphics[scale=0.6]{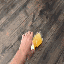}
&
\includegraphics[scale=0.6]{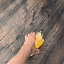}
&
\includegraphics[scale=0.6]{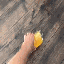}
&
\includegraphics[scale=0.6]{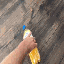}
&
\includegraphics[scale=0.6]{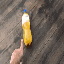}
\\
\end{tabular}
\\

\midrule

\begin{tabular}{l}
G: move book
\\
O: open handbag
\end{tabular}
&
\begin{tabular}{c}
\includegraphics[scale=0.6]{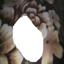}
\end{tabular}
&

\begin{tabular}{c @{\hspace{0.5mm}} c @{\hspace{0.5mm}} c @{\hspace{0.5mm}} c @{\hspace{0.5mm}} c}
\includegraphics[scale=0.6]{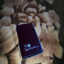}
&
\includegraphics[scale=0.6]{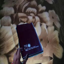}
&
\includegraphics[scale=0.6]{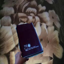}
&
\includegraphics[scale=0.6]{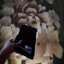}
&
\includegraphics[scale=0.6]{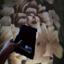}
\\
\end{tabular}
\\

\bottomrule
\end{tabular}
\caption{\textbf{Qualitative Evaluation (GS2).} Samples generated using our model in Generation Scenario 2, \textit{i.e.}, target action-object composition are unseen during training but target context image is seen with an object different from target object and a same/different action from target action. Thus, the overall target compositions comprising object, action and context are unseen during training. `G' indicates the target action-object composition and `O' indicates the action-object composition of the video (in the training set) from which the context image is chosen. We provide 5 frames for each generated video sample in the figure.}
\label{fig:gs2_samples} 
\end{figure}

\renewcommand{\arraystretch}{1}
\setlength{\tabcolsep}{7pt}
\begin{figure}[!h]
\centering
\scalebox{0.85}{
\begin{tabular}{c @{\hskip 0.04in } c @{\hskip 0.04in } c @{\hskip 0.04in } c @{\hskip 0.04in } c @{\hskip 0.04in } c @{\hskip 0.04in } c}
\toprule
\includegraphics[scale=0.8]{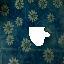}
&
\includegraphics[scale=0.8]{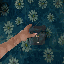}

&
\includegraphics[scale=0.8]{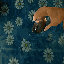}

&
\includegraphics[scale=0.8]{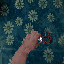}

&
\includegraphics[scale=0.8]{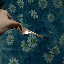}
&

\includegraphics[scale=0.8]{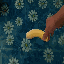}

&
\includegraphics[scale=0.8]{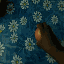}

\\

\footnotesize{O: push cup}
&
\footnotesize{G: lift \textit{handbag}}
&
\footnotesize{G: lift \textit{mouse}}
&
\footnotesize{G: lift \textit{scissors}}
&
\footnotesize{G: lift \textit{spoon}}

&
\footnotesize{G: lift \textit{banana}}
&
\footnotesize{G: lift \textit{apple}}
\\



\\
\midrule
\includegraphics[scale=0.8]{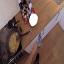}
&
\includegraphics[scale=0.8]{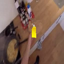}
&
\includegraphics[scale=0.8]{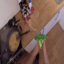}
&
\includegraphics[scale=0.8]{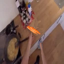}
&
\includegraphics[scale=0.8]{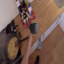}
&
\includegraphics[scale=0.8]{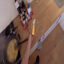}
&
\includegraphics[scale=0.8]{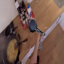}
\\
\footnotesize{O: take bowl}
& 
\footnotesize{G: take \textit{bottle}}
&
\footnotesize{G: take \textit{broccoli}}
&
\footnotesize{G: take \textit{carrot}}
&
\footnotesize{G: take \textit{cup}}
&
\footnotesize{G: take \textit{fork}}
&
\footnotesize{G: take \textit{pan}}
\\
\\
\bottomrule
\end{tabular}
}
\caption{\textbf{Qualitative Evaluation (GS2 - same action, same context, different objects). }Samples generated using HOI-GAN in Generation Scenario 2 corresponding to a set of compositions with same context frame, same action and different objects. `G' indicates the target action-object composition and `O' indicates the action-object composition of the video (in the training set) from which the context image is chosen. We show the context frame with mask on the left in each row. We provide 1 frame for each generated video sample in the figure.}
\label{fig:gs2_ao_setting1}
\end{figure}

\renewcommand{\arraystretch}{1}
\setlength{\tabcolsep}{7pt}
\begin{figure}[!h]
\centering
\scalebox{0.85}{
\begin{tabular}{c @{\hskip 0.04in } c @{\hskip 0.04in } c @{\hskip 0.04in } c @{\hskip 0.04in } c @{\hskip 0.04in } c @{\hskip 0.04in } c}
\toprule
\includegraphics[scale=0.8]{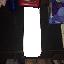}
&
\includegraphics[scale=0.8]{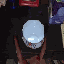}

&
\includegraphics[scale=0.8]{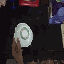}

&
\includegraphics[scale=0.8]{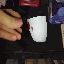}

&
\includegraphics[scale=0.8]{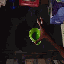}

&
\includegraphics[scale=0.8]{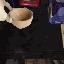}

&
\includegraphics[scale=0.8]{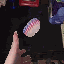}
\\
\footnotesize{O: open book}
&
\footnotesize{G: \textit{hold} bowl}
&
\footnotesize{G: \textit{move} bowl}
&
\footnotesize{G: \textit{push} bowl}
&
\footnotesize{G: \textit{put} bowl}
&
\footnotesize{G: \textit{remove} bowl}
&
\footnotesize{G: \textit{throw} bowl}
\\
\bottomrule
\end{tabular}
}

\caption{\textbf{Qualitative Evaluation (GS2 - same object, same context, different actions). } Samples generated using HOI-GAN in Generation Scenario 2 corresponding to a set of compositions with same context frame, same object and different actions. `G' indicates the target action-object composition and `O' indicates the action-object composition of the video (in the training set) from which the context image is chosen. We show the context frame with mask on the left in each row. We provide 1 frame for each generated video sample in the figure. }
\label{fig:gs2_ao_setting2}

\end{figure}

\vspace{4mm}
\noindent
\textbf{Qualitative Evaluation (GS1).} We present samples generated using our HOI-GAN in generaton scenario 1 (GS1). In GS1 setting, the target context image and the target action-object composition are unseen during training. Thus, the context image is from the test set (obtained in zero-shot compositional setting) and the object mask in the context image corresponds to the target object. As shown in Figure~\ref{fig:gs1_samples}, our model is able to create objects and enact the prescribed action on the object in the given context. 
Figure~\ref{fig:gs1_samples} also shows the real videos from the test set corresponding to the given compositions and context frame. The results clearly demonstrate that our model is able to generate realistic videos depicting the given action-object in the given context. 
The visual appearance of objects and actions (hand movements) are somewhat different in the generated videos compared to the corresponding real video 
because the model had to generalize based on other depictions of the object and action that were seen separately in training. 
Nevertheless, the results show that the generated video is also a realistic depiction of the target composition showing the target action on the target object in the target context. 

\renewcommand{\arraystretch}{0.8}
\setlength{\tabcolsep}{4pt}
\begin{figure}[!ht]
\centering
\scalebox{0.95}{
\begin{tabular}{c}
\toprule
\begin{tabular}{ccc}
$(a,o)$ & Context & Generated Video \\
\cmidrule(lr){1-1}\cmidrule(lr){2-2}\cmidrule(lr){3-3}
\begin{tabular}{c}target: lift handbag\end{tabular} 
&
\begin{tabular}{c}\includegraphics[scale=0.5]{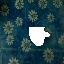}\end{tabular}
&
\begin{tabular}{ccccc}
\includegraphics[scale=0.5]{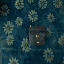}
&
\includegraphics[scale=0.5]{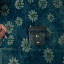}
&
\includegraphics[scale=0.5]{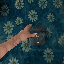}
&
\includegraphics[scale=0.5]{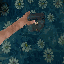}
&
\includegraphics[scale=0.5]{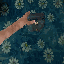}
\end{tabular}
\\
\midrule
\end{tabular}

\\

\begin{tabular}{c c}
training samples: action $a$ & training samples: object $o$ \\

\begin{tabular}{c}
training: lift apple
\\
\begin{tabular}{cccc}
\includegraphics[scale=0.5]{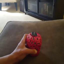}
&
\includegraphics[scale=0.5]{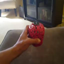}
&
\includegraphics[scale=0.5]{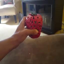}
&
\includegraphics[scale=0.5]{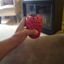}
\end{tabular}
\\
training: lift keyboard
\\
\begin{tabular}{cccc}
\includegraphics[scale=0.5]{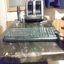}
&
\includegraphics[scale=0.5]{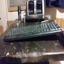}
&
\includegraphics[scale=0.5]{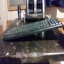}
&
\includegraphics[scale=0.5]{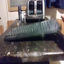}
\end{tabular}
\\
\end{tabular}

&

\begin{tabular}{c}
training: open handbag
\\
\begin{tabular}{cccc}
\includegraphics[scale=0.5]{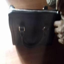}
&
\includegraphics[scale=0.5]{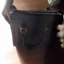}
&
\includegraphics[scale=0.5]{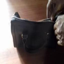}
&
\includegraphics[scale=0.5]{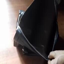}
\end{tabular}
\\
training: put handbag
\\
\begin{tabular}{cccc}
\includegraphics[scale=0.5]{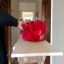}
&
\includegraphics[scale=0.5]{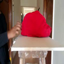}
&
\includegraphics[scale=0.5]{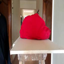}
&
\includegraphics[scale=0.5]{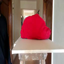}
\end{tabular}
\\
\end{tabular}
\\
\end{tabular}

\\
\\

\begin{tabular}{ccc}
\toprule
$(a,o)$ & Context & Generated Video \\
\cmidrule(lr){1-1}\cmidrule(lr){2-2}\cmidrule(lr){3-3}
\begin{tabular}{c}target: take pizza\end{tabular} 
&
\begin{tabular}{c}\includegraphics[scale=0.5]{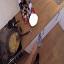}\end{tabular}
&
\begin{tabular}{ccccc}
\includegraphics[scale=0.5]{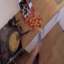}
&
\includegraphics[scale=0.5]{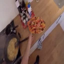}
&
\includegraphics[scale=0.5]{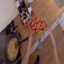}
&
\includegraphics[scale=0.5]{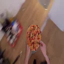}
&
\includegraphics[scale=0.5]{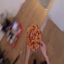}
\end{tabular}
\\
\midrule
\end{tabular}

\\
\begin{tabular}{c c}
training samples: action $a$ & training samples: object $o$ \\

\begin{tabular}{c}
training: take bottle
\\

\begin{tabular}{cccc}
\includegraphics[scale=0.5]{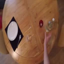}
&
\includegraphics[scale=0.5]{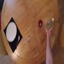}
&
\includegraphics[scale=0.5]{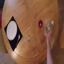}
&
\includegraphics[scale=0.5]{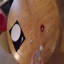}
\end{tabular}
\\
training: take sandwich
\\
\begin{tabular}{cccc}
\includegraphics[scale=0.5]{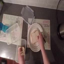}
&
\includegraphics[scale=0.5]{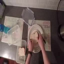}
&
\includegraphics[scale=0.5]{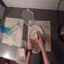}
&
\includegraphics[scale=0.5]{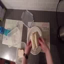}
\end{tabular}
\\
\end{tabular}

& 

\begin{tabular}{c}
training: put pizza
\\
\begin{tabular}{cccc}
\includegraphics[scale=0.5]{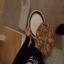}
&
\includegraphics[scale=0.5]{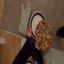}
&
\includegraphics[scale=0.5]{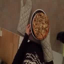}
&
\includegraphics[scale=0.5]{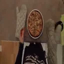}
\end{tabular}
\\
training: cut pizza
\\
\begin{tabular}{cccc}
\includegraphics[scale=0.5]{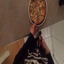}
&
\includegraphics[scale=0.5]{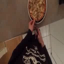}
&
\includegraphics[scale=0.5]{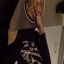}
&
\includegraphics[scale=0.5]{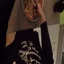}
\end{tabular}
\\
\end{tabular}

\\
\end{tabular}

\\
\\

\begin{tabular}{ccc}
\toprule
$(a,o)$ & Context & Generated Video \\
\cmidrule(lr){1-1}\cmidrule(lr){2-2}\cmidrule(lr){3-3}
\begin{tabular}{c}target: move broccoli \end{tabular} 
&
\begin{tabular}{c}\includegraphics[scale=0.5]{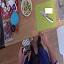}\end{tabular}
&
\begin{tabular}{ccccc}
\includegraphics[scale=0.5]{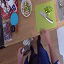}
&
\includegraphics[scale=0.5]{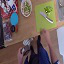}
&
\includegraphics[scale=0.5]{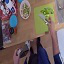}
&
\includegraphics[scale=0.5]{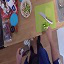}
&
\includegraphics[scale=0.5]{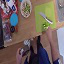}
\end{tabular}
\\
\midrule
\end{tabular}

\\
\begin{tabular}{c c}
training samples: action $a$ & training samples: object $o$ \\

\begin{tabular}{c}
training: move plate
\\

\begin{tabular}{cccc}
\includegraphics[scale=0.5]{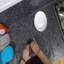}
&
\includegraphics[scale=0.5]{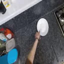}
&
\includegraphics[scale=0.5]{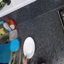}
&
\includegraphics[scale=0.5]{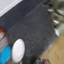}
\end{tabular}
\\
training: move pan
\\
\begin{tabular}{cccc}
\includegraphics[scale=0.5]{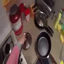}
&
\includegraphics[scale=0.5]{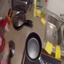}
&
\includegraphics[scale=0.5]{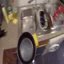}
&
\includegraphics[scale=0.5]{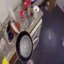}
\end{tabular}
\\
\end{tabular}

& 

\begin{tabular}{c}
training: wash broccoli
\\
\begin{tabular}{cccc}
\includegraphics[scale=0.5]{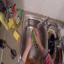}
&
\includegraphics[scale=0.5]{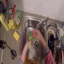}
&
\includegraphics[scale=0.5]{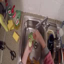}
&
\includegraphics[scale=0.5]{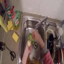}
\end{tabular}
\\
training: put broccoli
\\
\begin{tabular}{cccc}
\includegraphics[scale=0.5]{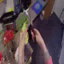}
&
\includegraphics[scale=0.5]{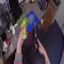}
&
\includegraphics[scale=0.5]{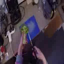}
&
\includegraphics[scale=0.5]{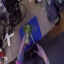}
\end{tabular}
\\
\end{tabular}

\\
\end{tabular}
\\
\bottomrule
\end{tabular}
}
\caption{\textbf{How does HOI-GAN generalize over compositions?.} Training samples in the data to illustrate that HOI-GAN leverages the information available during training and learns to combine them in a meaningful way. This ability allows HOI-GAN to generalize over unseen compositions of action, object and context. We provide a few frames for each sample in the figure.}
\label{fig:analysis}
\end{figure}

\renewcommand{\arraystretch}{1}
\setlength{\tabcolsep}{4pt}
\begin{figure}[!ht]
\centering
\scalebox{0.95}{
\begin{tabular}{p{1.2cm} c c c c c}
\toprule
$(a,o)$ & Context & C-VGAN & C-TGAN & MoCoGAN & HOI-GAN \\
\midrule
\begin{tabular}{p{1cm}} lift fork\end{tabular}
& 
\begin{tabular}{c} \includegraphics[scale=0.6]{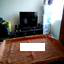}\end{tabular}
&
\begin{tabular}{c} \includegraphics[scale=0.6]{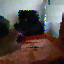}\end{tabular}
&
\begin{tabular}{c} \includegraphics[scale=0.6]{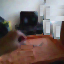}\end{tabular}
&
\begin{tabular}{c} \includegraphics[scale=0.6]{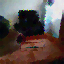}\end{tabular}
&
\begin{tabular}{c} \includegraphics[scale=0.6]{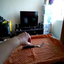}\end{tabular}
\\
\begin{tabular}{p{1cm}} bend carrot\end{tabular}
& 
\begin{tabular}{c} \includegraphics[scale=0.6]{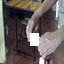}\end{tabular}
&
\begin{tabular}{c} \includegraphics[scale=0.6]{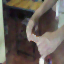}\end{tabular}
&
\begin{tabular}{c} \includegraphics[scale=0.6]{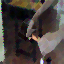}\end{tabular}
&
\begin{tabular}{c} \includegraphics[scale=0.6]{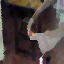}\end{tabular}
&
\begin{tabular}{c} \includegraphics[scale=0.6]{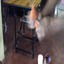}\end{tabular}
\\
\begin{tabular}{p{1cm}} put spoon\end{tabular}
& 
\begin{tabular}{c} \includegraphics[scale=0.6]{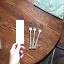}\end{tabular}
&
\begin{tabular}{c} \includegraphics[scale=0.6]{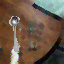}\end{tabular}
&
\begin{tabular}{c} \includegraphics[scale=0.6]{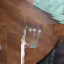}\end{tabular}
&
\begin{tabular}{c} \includegraphics[scale=0.6]{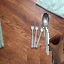}\end{tabular}
&
\begin{tabular}{c} \includegraphics[scale=0.6]{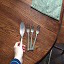}\end{tabular}
\\
\begin{tabular}{p{1cm}} open lid\end{tabular}
& 
\begin{tabular}{c} \includegraphics[scale=0.6]{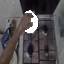}\end{tabular}
&
\begin{tabular}{c} \includegraphics[scale=0.6]{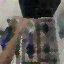}\end{tabular}
&
\begin{tabular}{c} \includegraphics[scale=0.6]{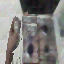}\end{tabular}
&
\begin{tabular}{c} \includegraphics[scale=0.6]{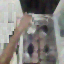}\end{tabular}
&
\begin{tabular}{c} \includegraphics[scale=0.6]{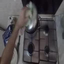}\end{tabular}
\\

\begin{tabular}{p{1.1cm}}cover banana\end{tabular}
& 
\begin{tabular}{c} \includegraphics[scale=0.6]{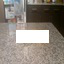}\end{tabular}
&
\begin{tabular}{c} \includegraphics[scale=0.6]{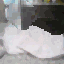}\end{tabular}
&
\begin{tabular}{c} \includegraphics[scale=0.6]{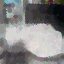}\end{tabular}
&
\begin{tabular}{c} \includegraphics[scale=0.6]{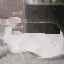}\end{tabular}
&
\begin{tabular}{c} \includegraphics[scale=0.6]{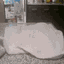}\end{tabular}
\\

\begin{tabular}{p{1cm}} fall ~~ carrot\end{tabular}
& 
\begin{tabular}{c} \includegraphics[scale=0.6]{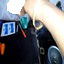}\end{tabular}
&
\begin{tabular}{c} \includegraphics[scale=0.6]{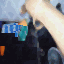}\end{tabular}
&
\begin{tabular}{c} \includegraphics[scale=0.6]{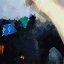}\end{tabular}
&
\begin{tabular}{c} \includegraphics[scale=0.6]{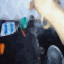}\end{tabular}
&
\begin{tabular}{c} \includegraphics[scale=0.6]{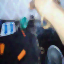}\end{tabular}
\\

\begin{tabular}{p{1cm}} brush pan\end{tabular}
& 
\begin{tabular}{c} \includegraphics[scale=0.6]{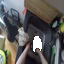}\end{tabular}
&
\begin{tabular}{c} \includegraphics[scale=0.6]{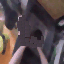}\end{tabular}
&
\begin{tabular}{c} \includegraphics[scale=0.6]{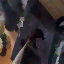}\end{tabular}
&
\begin{tabular}{c} \includegraphics[scale=0.6]{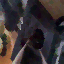}\end{tabular}
&
\begin{tabular}{c} \includegraphics[scale=0.6]{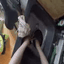}\end{tabular}
\\

\begin{tabular}{p{1cm}} fold cloth\end{tabular}
& 
\begin{tabular}{c} \includegraphics[scale=0.6]{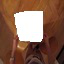}\end{tabular}
&
\begin{tabular}{c} \includegraphics[scale=0.6]{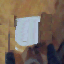}\end{tabular}
&
\begin{tabular}{c} \includegraphics[scale=0.6]{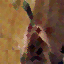}\end{tabular}
&
\begin{tabular}{c} \includegraphics[scale=0.6]{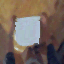}\end{tabular}
&
\begin{tabular}{c} \includegraphics[scale=0.6]{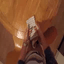}\end{tabular}
\\

\bottomrule
\end{tabular}
}
\caption{\textbf{Qualitative Evaluation (Baselines).} Samples generated using the baseline models (C-VGAN, C-TGAN, and MoCoGAN) in different generation scenarios. We also present the sample generated using HOI-GAN in the given composition of action-object $(a,o)$ pair and context image for comparison. We provide 1 frame for each generated video sample in the figure.}
\label{fig:baselines}
\end{figure}

\vspace{4mm}
\noindent
\textbf{Qualitative Evaluation (GS2).} We present samples generated using our HOI-GAN in generation scenario 2 (GS2). In GS2 setting, the target context background is seen during training while the target action-object composition is unseen. Specifically, the context image is from a video in the training set and the object mask in the context image corresponds to an object different than the target object. Also, the action corresponding to the context image may or may not be same as the target action. 
As such, the background may or may not be fully amenable for the target action-object composition.
As shown in Figure~\ref{fig:gs2_samples}, our model is able to create the required objects and enact the prescribed actions on the objects in the given context background. Moreover, our model is also able to modify the background as and when needed based on the target composition to be generated. The results clearly demonstrate that our model is able to generate realistic videos depicting the given action-object in the given context.
In particular, the \textit{move book} sample provides a comparison with its corresponding sample of \textit{move book} in the GS1 setting (see Figure~\ref{fig:gs1_samples}). In the GS2 setting seen here, the mask in the context frame corresponds to a handbag. The model is able to align the orientation of the book with the provided mask of the handbag and fit the object \textit{book} in the mask. In contrast, the size of \textit{book} with respect to the mask in this case is different from that seen in the GS1 example. 

In addition to showing the diversity in generated samples, we also generate videos corresponding to various sets of compositions with the same target context, same target action and different target objects. Samples in Figure~\ref{fig:gs2_ao_setting1} indicate our model is able to synthesize videos with the same action in the same context being performed on multiple objects differently. For instance, hand(s) appear from different directions and look different. Our model is also able to scale the objects appropriately based on the mask (see \textit{lift handbag} in Figure~\ref{fig:gs2_ao_setting1}). 

Furthermore, we also generate videos corresponding to various sets of compositions with the same target context, same target object and different target actions. Samples in Figure~\ref{fig:gs2_ao_setting2} indicate that our model is able to synthesize videos with different actions being performed on same object. In particular, the model is able to generate the same object with different and diverse set of visual appearances (\textit{e.g.} the \textit{bowl}s in Figure~\ref{fig:gs2_ao_setting2} look different) and perform the different actions upon them.  

\vspace{4mm}
\noindent
\textbf{How does HOI-GAN generalize over compositions?} Recall, the generation in this paper is performed in a zero-shot compositional setting, i.e., actions and objects are seen independently in certain compositions during training but the target action-object compositions are unseen during training. 
Intuitively, during this process, our model is able to effectively disentangle actions and objects.  
Therefore, when given a previously unseen target action-object composition for generation, our model is able to bring together or combine the information (distributed over the training set) in a meaningful way to synthesize realistic videos corresponding to the unseen composition. 
Consider the video corresponding to \textit{lift handbag} in Figure~\ref{fig:analysis}, the model has seen different handbags in different contexts with different actions (other than \textit{lift}), and has also seen different instances of the action \textit{lift} being performed on objects other than \textit{handbag} in different contexts. 
Given all this information, our model is able to combine the relevant information and synthesizes a video corresponding to a handbag being lifted in the given context. 
Similarly, we show two other compositions and the corresponding training samples of the action and object that might have helped the model during the particular generations.

\vspace{4mm}
\noindent
\textbf{Failure Cases (Additional Discussion).} We showed two failure cases in Section 4.3. Particularly, for \textit{open microwave}, while the model is able to generate a microwave object having seen it in training, it is not able to blend it into the background context. This is because the mask covers most of the background and the model gets very little information about the context. In the case of \textit{cut peach}, the model is unable to generate the pieces well because the interior of a peach differs from its exterior. This is in contrast to \textit{cut carrot} (see Figure~\ref{fig:gs2_samples}) wherein the interior of the carrot is similar to its exterior, and hence the model is able to generate the pieces properly.

\subsection{Qualitative Evaluation of Baselines}
In this section, we provide the middle frame of samples generated using the baselines: C-VGAN, C-TGAN, and MoCoGAN  for a given composition of context frame, action and object as conditional inputs. Figure~\ref{fig:baselines} shows the samples generated from these baselines. Figure~\ref{fig:baselines} also shows the samples generated using HOI-GAN corresponding to the given composition for comparison. 
The results clearly show that our HOI-GAN is able to synthesize more realistic videos. Moreover, this also supports the quantitative evaluation conducted in the main paper. Please view the samples together in the video on this webpage \url{http://www.sfu.ca/~mnawhal/projects/zs_hoi_generation.html}.

\setlength{\tabcolsep}{4.5pt}
\renewcommand{\arraystretch}{1}
\begin{table*}[t]
\centering
\caption{\textbf{Quantitative Evaluation (Effect of Word Embeddings).} Comparison of HOI-GAN with C-VGAN, C-TGAN, and MoCoGAN baselines using one-hot encoded labels instead of embeddings as conditional inputs(default version). (see section~\ref{sec:results}). Arrows indicate whether lower ($\downarrow$) or higher ($\uparrow$) is better. [I: inception score; S: saliency score; D: diversity score]}
\scalebox{0.9}{
\begin{tabular}{@{\hspace{1mm}}l@{\hspace{0.5mm}}l@{\hspace{0.1mm}}c@{\hspace{0.0mm}}c@{\hspace{0.0mm}}c@{\hspace{0.0mm}}c}
\toprule
&\multirow{3}{*}{Model} & \multicolumn{2}{c}{EPIC} & \multicolumn{2}{c}{SS} \\
\cmidrule(lr){3-4}\cmidrule(lr){5-6}
& & GS1 & GS2 & GS1 & GS2 \\
\cmidrule(lr){3-3}\cmidrule(lr){4-4}\cmidrule(lr){5-5}\cmidrule(lr){6-6}
& & 
I$\uparrow$ \hspace{1mm} S$\downarrow$ \hspace{1.2mm} D$\uparrow$
& 
I$\uparrow$ \hspace{1mm} S$\downarrow$ \hspace{1.2mm} D$\uparrow$
&
I$\uparrow$ \hspace{1mm} S$\downarrow$ \hspace{1.2mm} D$\uparrow$
& 
I$\uparrow$ \hspace{1mm} S$\downarrow$ \hspace{1.2mm} D$\uparrow$
\\
\midrule
& \small C-VGAN~\cite{vondrick2016generating}
&
\begin{tabular}{c c c}
1.1 & 52.1 & 0.4 
\end{tabular}
&
\begin{tabular}{c c c}
1.1 & 52.1 & 0.4 
\\ 
\end{tabular}
&
\begin{tabular}{c c c}
2.1 & 45.6 & 0.8 
\end{tabular}
&
\begin{tabular}{c c c}
1.9 & 45.1 & 0.5
\\ 
\end{tabular}
\\
& \small C-TGAN~\cite{saito2017temporal}
&
\begin{tabular}{c c c}
1.6 & 65.4 & 0.4 
\end{tabular}
&
\begin{tabular}{c c c}
2.2 & 28.1 & 0.5
\\ 
\end{tabular}
&
\begin{tabular}{c c c}
2.4 & 36.2 & 1.1
\end{tabular}
&
\begin{tabular}{c c c}
1.7 & 42.8 & 0.6
\\ 
\end{tabular}
\\
& \small MoCoGAN~\cite{tulyakov2018mocogan}
&
\begin{tabular}{c c c}
2.6 & 25.4 & 1.0 
\end{tabular}
&
\begin{tabular}{c c c}
2.0 & 34.9 & 1.0 
\end{tabular}
&
\begin{tabular}{c c c}
2.9 & 22.8 & 1.3
\end{tabular}
&
\begin{tabular}{c c c}
2.4 & 27.4 & 1.5
\end{tabular}
\\
\midrule
\multirow{2}{*}{\rotatebox[origin=c]{90}{{\footnotesize (ours)}}}
& HOI-GAN (bboxes)
&
\begin{tabular}{c c c}
3.8 & 18.5 & 2.1\\ 
\end{tabular}
&
\begin{tabular}{c c c}
3.2 & 24.1 & 2.4\\ 
\end{tabular}
&
\begin{tabular}{c c c}
4.9 & 26.2 & 2.7\\
\end{tabular}
&
\begin{tabular}{c c c}
4.0& 25.2&2.4\\
\end{tabular}
\\
& HOI-GAN (masks)
&
\begin{tabular}{c c c}
\hspace{0.8mm}\textbf{4.3} & \hspace{-0.8mm}\textbf{16.5} & \hspace{-0.8mm}\textbf{2.5}\\ 
\end{tabular}
&
\begin{tabular}{c c c}
\hspace{0.8mm}\textbf{3.9} & \hspace{-0.8mm}\textbf{20.2} & \hspace{-0.8mm}\textbf{1.6}\\ 
\end{tabular}
&
\begin{tabular}{c c c}
\hspace{0.8mm}\textbf{5.8} & \hspace{-0.8mm}\textbf{15.8} & \hspace{-0.8mm}\textbf{3.0}
\end{tabular}
&
\begin{tabular}{c c c}
\hspace{0.8mm}\textbf{4.5} & \hspace{-0.8mm}\textbf{23.7} & \hspace{-0.8mm}\textbf{2.8}
\end{tabular}
\\
\bottomrule
\end{tabular}
}

\label{tab:combined_labVsemb}
\end{table*}
\clearpage
\subsection{Additional Quantitative Evaluation}
In this section, we provide results of the additional quantitative evaluation of our HOI-GAN to illustrate the effect of using semantic embeddings.

\subsubsection{Effect of Word Embeddings.} 
In our approach, we use word embeddings for the action and object labels to share information among semantically similar categories during training. To demonstrate the impact of using embeddings, we also trained HOI-GAN using one-hot encoded labels corresponding to both actions and objects. We observe that these models perform worse than the models trained using semantic embeddings (refer last two rows of Table~\ref{tab:combined} in the main paper and Table~\ref{tab:combined_labVsemb}). Nevertheless, our models still outperform the baselines (refer to Table~\ref{tab:combined_labVsemb}).

\begin{figure*}[b]
\centering
    \begin{tabular}{c}
         \includegraphics[width=0.9\textwidth]{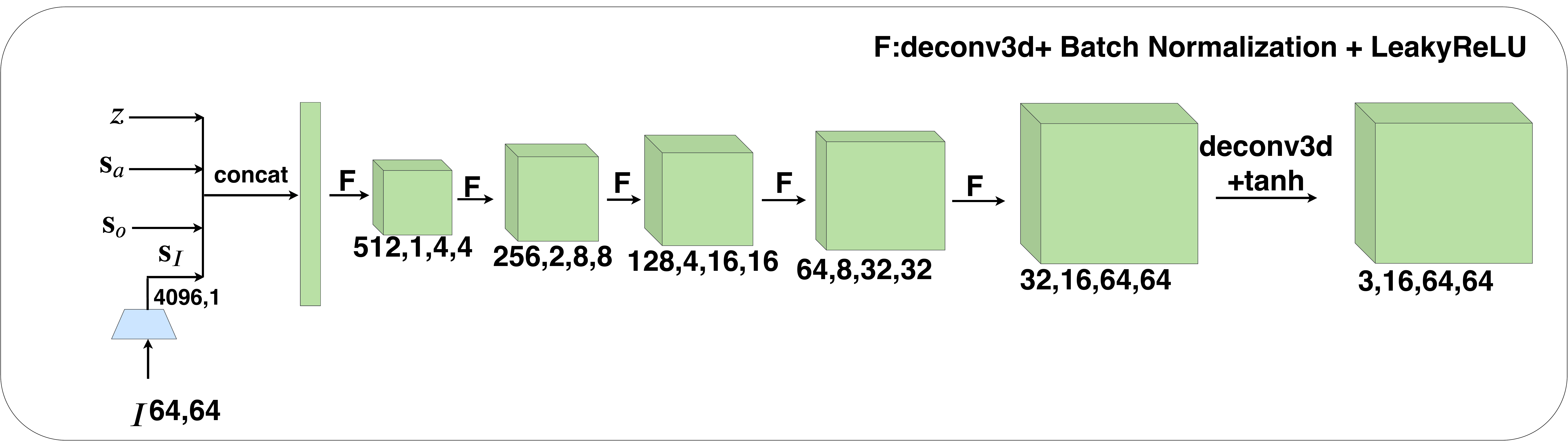}\\
         \textbf{(i)} Generator Network in HOI-GAN\\\\\\
         \includegraphics[width=0.9\textwidth]{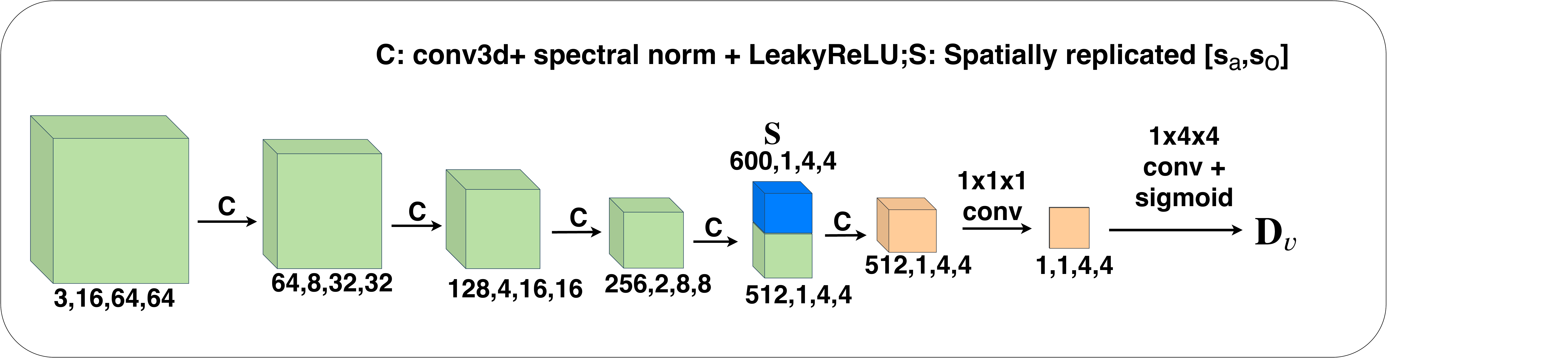}\\
         \textbf{(ii)} Video Discriminator Network in HOI-GAN\\\\\\
         \includegraphics[width=0.9\textwidth]{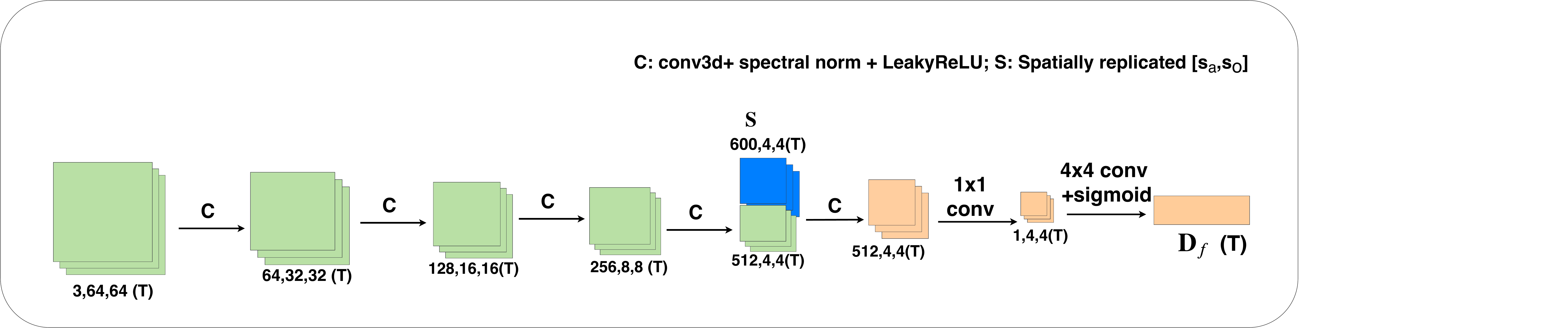}\\
         \textbf{(iii)} Frame Discriminator Network in HOI-GAN\\\\\\
         \includegraphics[width=0.9\textwidth]{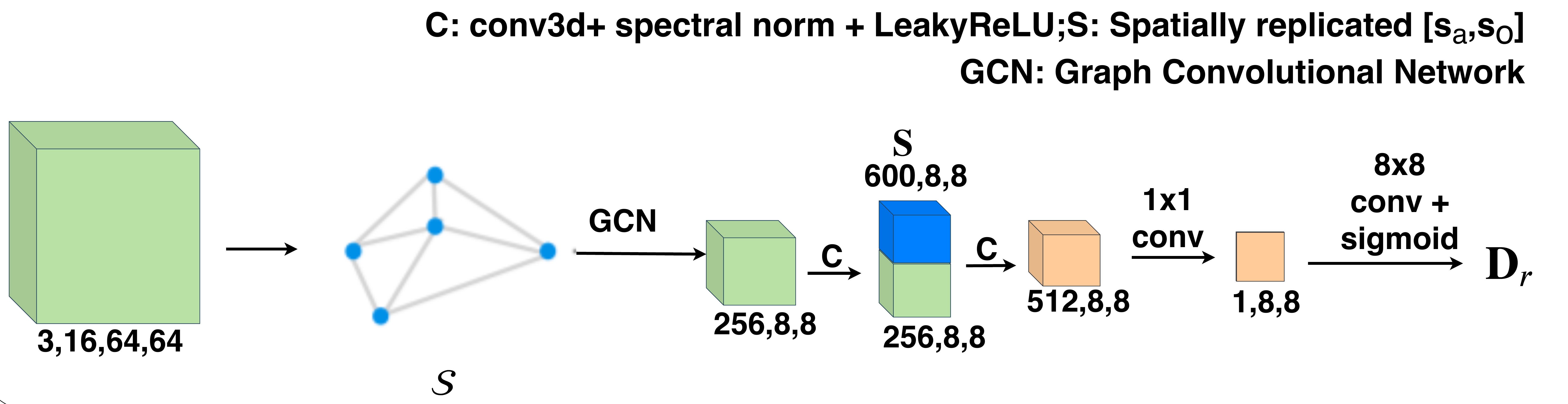}\\
         \textbf{(iv)} Relational Discriminator Network in HOI-GAN\\\\\\
    \end{tabular}
\caption{\textbf{Architecture Details.} Model architectures used in our experiments for: (i) Generator, (ii)Video Discriminator, (iii) Frame discriminator (gradient discriminator has similar architecture), (iv) Relational Discriminator. Best viewed in color on desktop.}
\label{fig:model_individual}
\end{figure*}

\subsubsection{Evaluation using FID}
\setlength{\tabcolsep}{8pt}
\renewcommand{\arraystretch}{1}
\begin{table*}[t]
\centering
\caption{\textbf{Quantitative Evaluation (FID).} Fr\'echet Inception Distance (FID) comparison of HOI-GAN with C-VGAN, C-TGAN, and MoCoGAN baselines. Lower FID implies higher quality. }
\begin{tabular}{lcccc}
\toprule
Model & \multicolumn{2}{c}{EPIC} & \multicolumn{2}{c}{SS} \\
\cmidrule(lr){1-1}\cmidrule{2-3}\cmidrule(lr){4-5}
& GS1 & GS2 & GS1 & GS2 
\\
\midrule
\small C-VGAN~\cite{vondrick2016generating} & 18.8 & 23.7 & 15.1 & 20.5
\\
\small C-TGAN~\cite{saito2017temporal} & 17.2 & 21.3 & 13.6 & 18.2 
\\
\small MoCoGAN~\cite{tulyakov2018mocogan} & 14.6 & 19.9 & 11.4 & 17.5
\\
\small HOI-GAN (ours) & \textbf{8.1} & \textbf{10.2} & \textbf{7.2} & \textbf{8.3}\\
\bottomrule
\end{tabular}
\label{tab:fid}
\end{table*}

We primarily used video classifier based Inception score as a metric for quantitative evaluation. As an additional measure to evaluate the quality of generated samples, we also report another Fr\'echet Inception Distance (lower is better) in Table~\ref{tab:fid}. We compute the scores following ~\cite{wang2018vid2vid}. Specifically, we use a Kinetics-pretrained ResNext-101 video classification model as the feature extractor. The results show that videos generated using HOI-GAN are more realistic than those created using baselines.

\subsubsection{Classification Experiments} To further demonstrate the effectiveness of our model, we conduct classification experiments using generated videos in different settings. The experiments are described as follows.
\setlength{\tabcolsep}{4.7pt}
\renewcommand{\arraystretch}{1.0}
\begin{table}[!t]
\centering
\caption{\textbf{Classification Experiments.} Accuracy of a video classifier when finetuned on real videos from the dataset and evaluated on generated videos corresponding to unseen action-object compositions. 
}
\begin{tabular}{l@{\hskip 8mm}c@{\hskip 7mm}c}
\toprule
\centering
Classifier Setting & EPIC & SS\\
\midrule
Chance & $<$0.1 & $<$0.1\\
Finetuned on real / Evaluated on generated (MoCoGAN) & 11.0 & 20.6 \\
Finetuned on real / Evaluated on generated (HOI-GAN) & 35.4 & 53.6 \\
Finetuned on real / Evaluated on real & 51.7 & 68.8\\
\bottomrule
\end{tabular}
\label{tab:classification2}
\end{table}

\textit{Finetuning on real and evaluation on generated videos.} We finetuned a Kinetics-pretrained ResNext-101 classifier model (same as the one used to compute evaluation metrics). We used this finetuned video classifier to classify generated videos. We report the classification performance of the classifier in Table~\ref{tab:classification2}. The evaluation is done for the generated videos corresponding to the unseen compositions only.  For our HOI-GAN and baseline MoCoGAN, we calculated the accuracy on the videos generated by the models (with unseen compositions as conditional input). For comparison, we also report the classification performance on a test set containing real videos of the same compositions -- this serves as the upper bound. We observe that the performance on videos generated using HOI-GAN is considerably better than that on videos generated using MocoGAN (best performing baseline) and much closer to the performance on real videos.  This indicates that our proposed framework is consistently generating realistic videos conditioned on given action-object compositions.

\setlength{\tabcolsep}{4.7pt}
\renewcommand{\arraystretch}{1.0}
\begin{table}[!t]
\centering
\caption{\textbf{Classification Experiments.} Accuracy of a video classifier when finetuned on generated videos and evaluated on real videos for unseen action-object compositions.}
\begin{tabular}{l@{\hskip 8mm}c@{\hskip 7mm}c}
\toprule
\centering
Classifier Setting & EPIC & SS\\
\midrule
Finetuned on generated / Evaluated on real & 33.1 & 46.3 \\
Finetuned on real / Evaluated on real & 51.7 & 68.8\\
\bottomrule
\end{tabular}
\label{tab:classification1}
\end{table}

\textit{Finetuning on generated and evaluation on real videos.} We used a Kinetics-pretrained ResNext-101 video classifier (same as the one used to compute evaluation metrics) and fine-tuned it on a dataset containing only generated samples corresponding to the unseen action-object compositions. We report the classification performance in terms of accuracy of this classifier when evaluated on a test set containing real videos corresponding to unseen compositions (from the original dataset) in Table~\ref{tab:classification1}. For reference, we also report the classification performance on the same test set for the classifier fine-tuned on real videos. As expected, performance is lower than that using real videos, but the generated ones serve as a reasonable proxy for learning to recognize unseen compositions.
\\

\subsection{Preprocessing and Data Splits}
As described in Section 4.1, we perform new splits of the dataset for the task of zero-shot HOI video generation. In this section, we provide the details of preprocessing and zero-shot compositional splits for datasets EPIC-Kitchens (EPIC) and 20BN-Something-Something V2 (SS).
\\

\vspace{0.06in}
\noindent
\textbf{EPIC: Processing and Splits.}
 The EPIC-Kitchens dataset originally consists of 39,594 video samples of the form $(V,a,o)$, \textit{i.e.}, video $V$ with action label $a$ and object label $o$, spanning 125 unique actions and 352 unique objects. We further filtered the dataset to ensure that the video samples contain both ground truth bounding box annotation and MaskRCNN output (NMS threshold = 0.7) in the frames uniformly sampled from a video. We interpolated the sequence if the number of such frames is less than 16. We then split the filtered dataset by action-object compositions to obtain train and test splits suitable for the zero-shot compositional setting, \textit{i.e.}, all the unique object and action labels in combined dataset appear independently in the train split, however, a certain pair of action and object present in the test split is absent in train split and vice versa. Subsequently we obtained two splits: (1) train split containing 19,895 videos that overall depict 1,128 unique action-object compositions, and (2) test split containing 7,805 videos (568 unique action-object compositions). The final splits consist of compositions spanning 204 unique actions and 63 unique objects.
\\

\vspace{0.06in}
\noindent
\textbf{SS: Processing and Splits.}
 The 20BN-Something-Something V2 dataset originally consists of 220,847 video samples of the form $(V,l)$, \textit{i.e.}, video $V$ having a label $l$. To transform the dataset instances to the form $(V,a,o)$, we applied NLTK POS-tagger on $l$ and obtained verb $a$ and noun $o$. In particular, we considered the verb tag (after stemming) in $l$ as action label $a$. We observe that all instances of $l$ begin with the present continuous form of $a$ which is acting upon the subsequent noun. Therefore, we used the noun that appears immediately after the verb as object ${o}$. We merged the train and validation split of the transformed dataset. We further filtered the dataset to ensure that the video samples contain objects that can be detected using MaskRCNN (NMS threshold = 0.7) in the frames uniformly sampled from a video. We then split the transformed dataset by compositions of action $a$ and object $o$ to obtain the train and test splits suitable for the zero-shot compositional setting (same as EPIC). Subsequently, we obtained two splits: (1) train split containing 23,511 videos overall that overall depict 671 unique action-object compositions, and (2) test split containing 3,515 videos overall (135 unique action-object compositions). The final splits consist of compositions spanning 48 unique actions and 62 unique objects.

\subsection{Implementation Details}
In our experiments, the convolutional layers in all networks, namely, $\mathbf{G}$, $\mathbf{D}_f$, $\mathbf{D}_g$, $\mathbf{D}_v$, $\mathbf{D}_r$ have kernel size 4 and stride 2. We generate a video clip consisting of $T = 16$ frames having $H =W = 64$. The noise vector $\mathbf{z}$ is of length 100. The parameters $w_0 = h_0 = 4$, $d_0 = 1$ and $N=512$ for $\mathbf{D}_v$ and $w^t_0 = h^t_0 = 4$ and $N^{(t)}=512$ for $\mathbf{D}_f$, $\mathbf{D}_g$, and $\mathbf{D}_r$. To obtain the semantic embeddings $\mathbf{s}_a$ and $\mathbf{s}_o$ corresponding to action and object labels respectively, we use Wikipedia-pretrained GLoVe \cite{pennington2014glove} embedding vectors of length 300. We provide further implementation details of our model architecture in the supplementary section. For training, we use the Adam \cite{kingma2015adam} optimizer with learning rate 0.0002 and $\beta_1 = 0.5$, $\beta_2= 0.999$. We train all our models with a batch size of 32. We use dropout (probability = 0.3) \cite{salimans2016improved} in the last layer of all discriminators and all layers (except first) of the generator.

\noindent
\textbf{Relational discriminator.} We used the final output layer of MaskRCNN, that comprises a list of bounding boxes, a list of segmentation masks and a list of labels corresponding to each detection. We used \url{https://github.com/facebookresearch/maskrcnn-benchmark} repository to obtain the detection output. The same list of bounding boxes have been used for real and generated. Then, using each bounding box in the output, we crop the visual region from the corresponding frame. These crops will correspond to the nodes of spatio-temporal graph. These cropped visual regions are resized to $3\times 16\times 16$ $(C\times H\times W)$ and their position (bounding box top-left coordinates normalized with respect to the image size) and their original aspect ratio (bounding box height and width normalized with respect to image size) are collectively used for node feature representation (Refer to Figure 3 for illustration). We used a \texttt{conv} module (shared weights for all crops), i.e., convolutional layers (stride=2, kernel size=4) and obtain a convolutional embedding for resized visual regions of size 4096 appended with 4 additional numbers corresponding to position and aspect ratio. We design Graph Convolution Layer using the implementation of Graph Convolution Network (GCN) available at \url{https://github.com/tkipf/pygcn}. We used 7 such Graph Convolution layers: initial layer converts the feature size to 4096 and output feature size of the node is doubled every two layer in next 6 layers. Until this stage, the node is represented using single dimensional vector. After pooling along the temporal axis, the channel dimension is reshaped to $256\times8\times8$ and the resulting tensor is of shape $K\times256\times8\times8$ where $K$ is the number of crops.

\subsubsection{Architecture Details.}
As described in Section 3, our model comprises 5 networks involving a generator network and four discriminator networks. We provide the details of the architectures used in our implementation for the generator network, video discriminator, frame discriminator and relational discriminator in Figure~\ref{fig:model_individual}. The architecture for gradient discriminator is same as that of the frame discriminator.

\end{document}